\pdfoutput=1
\documentclass[11pt]{article}

\usepackage[preprint]{acl}

\usepackage{times}
\usepackage{latexsym}

\usepackage{booktabs}
\usepackage{caption}
\usepackage{fullpage}
\usepackage{longtable}

\usepackage[T1]{fontenc}
\usepackage[utf8]{inputenc}

\usepackage{microtype}

\usepackage{inconsolata}

\usepackage{multirow}

\usepackage{graphicx}

\title{Prompt Design Matters for Computational Social Science Tasks \\ but in Unpredictable Ways}

\author{Shubham Atreja, Joshua Ashkinaze, Lingyao Li, {\bf Julia Mendelsohn}, {\bf Libby Hemphill} \\
 University of Michigan School of Information \\ \texttt{satreja@umich.edu}
  }

\begin{document}

\maketitle

\begin{abstract}
Manually annotating data for computational social science tasks can be costly, time-consuming, and emotionally draining. While recent work suggests that LLMs can perform such annotation tasks in zero-shot settings, little is known about how prompt design impacts LLMs' \emph{compliance} and \emph{accuracy}. We conduct a large-scale multi-prompt experiment to test how model selection (ChatGPT, PaLM2, and Falcon7b) and prompt design features (definition inclusion, output type, explanation, and prompt length) impact the compliance and accuracy of LLM-generated annotations on four CSS tasks (toxicity, sentiment, rumor stance, and news frames). Our results show that LLM compliance and accuracy are highly prompt-dependent. For instance, prompting for numerical scores instead of labels reduces all LLMs' compliance and accuracy. The overall best prompting setup is task-dependent, and minor prompt changes can cause large changes in the distribution of generated labels. By showing that prompt design significantly impacts the quality and distribution of LLM-generated annotations, this work serves as both a warning and practical guide for researchers and practitioners. 
\end{abstract}

\section{Introduction}

NLP systems for computational social science tasks have traditionally relied on manually annotating large datasets, which can yield high-quality labels but at the expense of time, money, and emotional labor. Many studies are thus turning to prompting LLMs for text annotations for many tasks such as toxicity \cite{li2024hot} and news frame detection \citep{gilardi2023chatgpt}. Results \cite{li2024hot, qin-etal-2023-chatgpt,gilardi2023chatgpt} show that LLMs like ChatGPT and PaLM can perform these text annotation tasks in \emph{zero-shot setting}, i.e., through prompts containing instructions on how to annotate the data. However, there is little large-scale, systematic, empirical evidence about what prompt designs are most effective across computational social science tasks. 

\begin{figure}
    \centering
    \includegraphics[width=0.7\columnwidth]{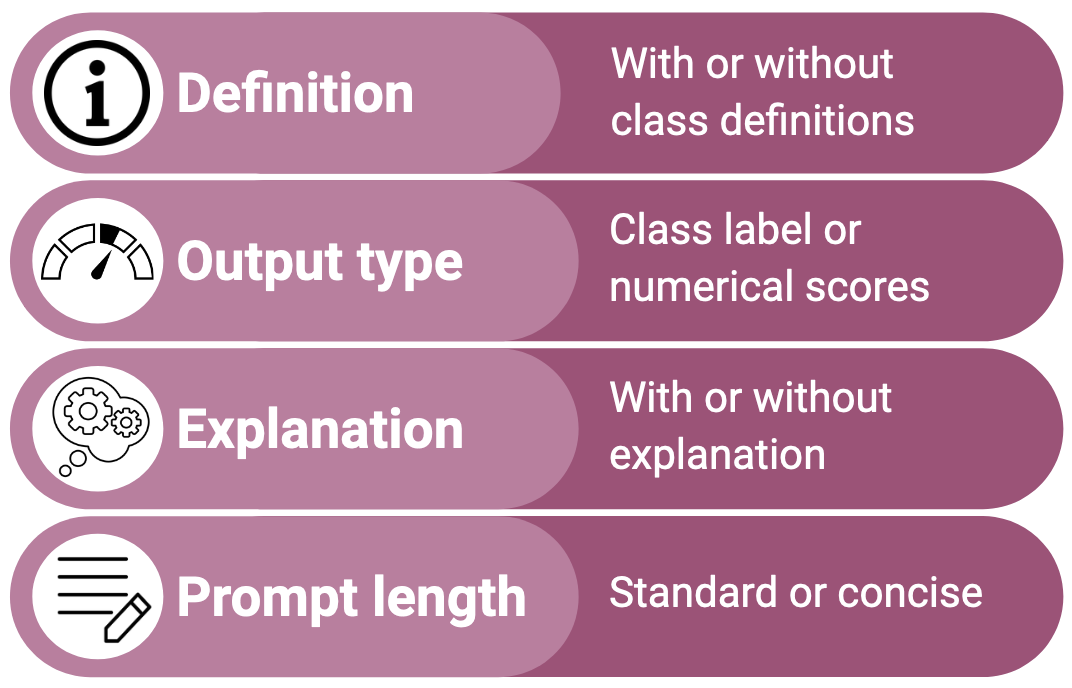}
    \caption{Prompt variations used in our experiments}
    \label{fig:prompt-design}
\end{figure}

Most research on benchmarking LLMs’ performance report results using just \textit{one prompt design} \cite{wang2023chatgpt,qin-etal-2023-chatgpt,gilardi2023chatgpt}. While numerous guides for LLM prompting exist \cite{diar2023ai, giray2023prompt, artprompting2023, bach2022promptsource}, they do not all offer the same guidance, and leave many empirical questions unanswered. For example, most guides suggest making the prompts as ``descriptive and detailed'' as possible \cite{diar2023ai}. However, longer prompts make tasks more expensive as LLM costs depend on the number of input tokens. Can re-writing prompts for concision still maintain accuracy? 

Separate from prompt designs that lead to accurate outputs, there is little systematic evidence on the extent of LLMs' \textit{compliance} with input prompt instructions. It is important that LLMs generate valid output that conforms to the instructions provided in the prompt since non-compliance wastes both time and money. \citet{qin-etal-2023-chatgpt} report some examples where ChatGPT does not comply with the input prompt -- despite explicit instructions to generate ``positive'' or ``negative'' sentiment labels only, ChatGPT generates ``neutral'' or ``mixed'' as the label. In the absence of any systemic evidence, however, it remains unclear whether certain prompt designs are more or less likely to generate compliant outputs. 

% HIGHLIGHTING OUR CONTRIBUTION

To understand the relationship between prompt design and LLM compliance and accuracy, we conducted a large-scale multi-prompt experiment to annotate four datasets including toxicity, sentiment, rumor stance, and news frames, using three LLMs (ChatGPT, PaLM2, Falcon7b). Inspired by a combination of popular prompting practices \cite{diar2023ai} and practical constraints (e.g., prompting costs), we vary prompts along four dimensions (see Figure \ref{fig:prompt-design}): i) definition inclusion (yes/no), ii) output type (label or numerical score), iii) explanation (yes/no), and iv) prompt length (standard/concise). We follow a complete factorial design to generate 16 different prompts (2*2*2*2) for each task and produce a large multi-task, multi-model, multi-prompt design experiment with a combined 362,928 annotations.

% i) prompting with or without label definitions, ii) prompting for an output label or numerical scores, iii) prompting for an explanation or not, and iv) varying the length of the prompt (standard or concise). 

% TAKEAWAYS BELOW THIS.... 

Our results show that LLM compliance and accuracy are highly prompt-dependent, especially for multi-class tasks, and that prompts' influences vary by model. For example, Falcon7b's compliance on rumor stance varies up to 55\% across different prompts. ChatGPT's accuracy on news framing varies up to 14\% across prompts. Below, we report our key findings for individual prompt designs:

\begin{itemize}
    \item Prompting for numerical scores instead of labels reduces both compliance and accuracy for most LLMs and tasks.

    \item Prompting with definitions improves ChatGPT's accuracy without reducing its compliance. Prompting with definitions reduces PaLM2's and Falcon7b's compliance. 

    \item The impact of concise prompts on accuracy and compliance is highly task and model dependent. For example, prompting PaLM2 with concise prompts reduces the cost of rumor stance annotations without decreasing compliance and accuracy. In most cases, however, concise prompts adversely impact either accuracy or compliance.

    \item Prompting LLMs to explain their input improves their compliance with prompt instructions. However, this also changes the distribution of generated labels. For example, ChatGPT annotates 34\% more content as neutral when prompted to explain its output. 
\end{itemize}

Taken together, we highlight inconsistent effects of prompt design features across tasks, but also point to several best practices for researchers and practitioners. Crucially, we caution that different prompting strategies can yield different annotation distributions which may in turn affect social science research results.

\section{Related Work}

\subsection{LLMs for NLP+CSS}

NLP systems for computational social science tasks often require manual annotations to train classifiers and evaluate the effectiveness of unsupervised models \cite{gilardi2023chatgpt}. In many instances, these applications demand the support of crowd-workers sourced from platforms such as MTurk to annotate data samples \cite{huynh2021survey, gilardi2023chatgpt}. However, the financial cost of data annotation is often high, and the demographics of annotators can influence the objectivity of the annotations \cite{diaz2022crowdworksheets}. For specific annotation tasks, such as toxicity detection, annotators are exposed to harmful and offensive content. This exposure limits the pool of available annotators and restricts the volume of content they can reasonably review \cite{li2024hot}.

The advancements of LLMs like ChatGPT and Google PaLM are transforming the landscape of annotation tasks in NLP \cite{gilardi2023chatgpt, kocon2023chatgpt}. A major advantage of using LLMs for annotations discussed extensively in the literature is the cost-effectiveness, as LLMs potentially offer a more economical solution for large-scale annotation needs \cite{wang2021want, gilardi2023chatgpt}. Moreover, as suggested by \citet{li2024hot}, using LLMs for such annotation can protect annotators, particularly those from marginalized groups, by sparing them from exposure to harmful content that could otherwise induce undue pressure \cite{li2024hot}. Another advantage of using LLMs for annotations is their explainability and reasoning capabilities \cite{zhang2022would, huang2023chatgpt, liu2023trustworthy}. Huang et al. (2023) observed that ChatGPT could generate quality explanations comparable to human annotators for implicit hate speech \cite{huang2023chatgpt}. 

Recent studies \cite{qin-etal-2023-chatgpt, kocon2023chatgpt} have presented substantial progress in using LLMs for annotations, which could help a broad set of NLP tasks, including but not limited to, sentiment classification \cite{wang2023chatgpt, okey2023investigating}, news summarization \cite{zhang2024benchmarking}, rumor detection \cite{liu2024can}, and toxicity identification \cite{li2024hot}. However, most research on benchmarking LLMs' performance on NLP tasks has reported results using just one prompt design. Even when researchers \cite{wang2023chatgpt, huang2023chatgpt} designed multiple prompts, they tested their prompts on a small sample and reported final results using only one prompt due to the high computational and monetary costs involved in testing complete datasets on multiple prompts.

\subsection{Prompt Design}

Prompts are a set of instructions designed to engage and guide the behavior of LLMs \cite{white2023prompt, giray2023prompt}. Typically, a prompt consists of four elements \cite{diar2023ai}: (1) Instruction – a specific task for the model to perform, (2) Context – additional information, such as concept definitions, to help generate better responses, (3) Input data – the question or data for the model to respond to or annotate, and (4) Output indicator – the desired type or format of the response. When designed properly, prompts can vastly expand the range of tasks that LLMs can handle without requiring new training data or modifications to the underlying models \cite{zhang2021differentiable}.

Researchers have explored a variety of prompting techniques to interact with LLMs, such as zero-shot \cite{xian2017zero}, few-shot \cite{brown2020language}, and chain-of-thought \cite{wei2022chain}. Amongst these, zero-shot prompting is most widely used as users can provide input instructions without needing additional labeled examples or training data \cite{wei2021finetuned}. Few-shot prompting is useful for in-context learning where LLMs can learn from a few input and output examples added in the prompt \cite{brown2020language, wang2020generalizing}. Chain-of-thought (CoT) prompting has recently gained attention due to its ability to elicit complex and multi-step reasoning by providing instructions in a step-by-step manner \cite{wei2022chain}. In our study, we use single-step zero-shot prompting as the approach is most scalable and easy to draft when annotating large datasets.

Numerous guides have also been released on formulating zero-shot input prompts \cite{diar2023ai, giray2023prompt, artprompting2023, bach2022promptsource}. Most guides suggest making the prompts as ``descriptive and detailed'' \cite{diar2023ai} as possible. Other guidelines include prompting with clear definitions to reduce the gap between humans and LLMs \cite{giray2023prompt, artprompting2023}. While generally useful, the guides provide little empirical evidence to back their claims. Empirically, \citet{li2024hot} introduced prompting for numerical scores to enhance LLM performance on toxicity detection by selecting different thresholds \cite{li2024hot}.\citet{nguyen2024human} introduced prompting for explanations and underlined LLMs' potential to generate human-like annotations.

\section{Experiment Details}
In this section, we first explain the different input prompts designed for our experiments and then describe the LLMs and tasks used for the experiment.

\subsection{Prompt Design} 

We limited our experiment to single-stage zero-shot prompts as they are the most cost-effective and scalable for annotating large datasets. First, we designed a prompt for each task and then introduced 4 variations (shown in Figure \ref{fig:prompt-design}) in each task-specific prompt. The prompts and their variations were inspired by a combination of prior work on prompting LLMs \cite{qin-etal-2023-chatgpt, ding-etal-2023-gpt, li2024hot} and practical factors such as the prompt's fixed annotation cost.

\textbf{Definition (yes or no):} prompting with or without output class definitions. We used the same definitions provided to human raters when the datasets were first annotated in prior work. We introduced this variation for all tasks except sentiment analysis as no sentiment definitions were made available in prior work. 

\textbf{Output type (label or score):} prompting for a final output label or numerical (probabilistic) scores for individual labels. \citet{li2024hot} introduced prompting for numerical scores to control the precision and recall in LLM-generated data. 

\textbf{Explanation (yes or no):} prompting the model to provide an explanation in its output or not. Explanations can add useful context to the LLM's performance and errors but it can also introduce challenges in automated parsing of the output. 

\textbf{Prompt length (standard or concise):} prompting with the standard prompt or its concise version. Standard prompts were descriptive and detailed to achieve best performance \cite{diar2023ai, artprompting2023}. Concise prompts were paraphrased versions (\string~ 53\% less words) of the standard prompt generated using GPT-3 to reduce the fixed cost per annotation as LLM API costs are dependent on the number of input tokens. We manually verified every concise prompt to ensure that they contain all the information from the standard prompt. 

More generally, each prompt variation is of a different length and can impact the fixed cost per annotation. Table \ref{tab:prompt-costs} shows the change in the number of words due to each prompt variation, which is indicative of the change in annotation costs. A list of all the prompts used in the experiment is provided in Appendix Table \ref{tab:all-prompts}.

\begin{table}[]
\resizebox{\columnwidth}{!}{
\begin{tabular}{@{}lc@{}}
\toprule
\textbf{Prompt design}      & \multicolumn{1}{l}{\textbf{$\Delta$(Num of words and fixed cost)}} \\ \midrule
Adding definitions          & {\color[HTML]{EA4335}+91.97\%}                                                    \\
Asking for explanation      & {\color[HTML]{EA4335}+10.31\%}                                                    \\
Asking for numerical scores & {\color[HTML]{EA4335}+22.37\%}                                                    \\
Concise version             & {\color[HTML]{34A853}-53.97\%}                                                    \\ \bottomrule
\end{tabular}
}
\caption{Changes in prompt length and fixed annotation cost due to different prompt designs}
\label{tab:prompt-costs}
\end{table}

\subsection{Models} 
We used three instruct-tuned LLMs in our experiment -- \textbf{ChatGPT (GPT3.5-turbo)}, \textbf{PaLM2 (chat-bison-001)}, and \textbf{Falcon7b-instruct}, which represent different architectures, sizes, and costs. \textbf{GPT3.5-turbo} is OpenAI's high performing inexpensive model shown to be effective at performing most NLP tasks \cite{gilardi2023chatgpt}. \textbf{PaLM2} is a family of generative models launched by Google and shown to outperform human raters on many tasks \cite{suzgun-etal-2023-challenging, sarkar-etal-2023-zero}. We picked \textbf{Falcon7b} as our third model to find out how smaller open source LLMs compare against larger models. Falcon7b is part of the Falcon series of open source models\footnote{https://falconllm.tii.ae/falcon.html} and has 7b parameters. Compared to its larger siblings, Falcon7b can be setup without a GPU. At the start of this study (June 2023), Falcon series was ranked highest on Hugging Face’s open source LLM leaderboard\footnote{https://huggingface.co/spaces/HuggingFaceH4/
open\_llm\_leaderboard}.

\subsection{Annotation Datasets and Tasks}
We conducted our experiments with 4 diverse tasks (see Table \ref{tab:dataset}) representing different numbers of output classes and different levels of complexity:

\textbf{Toxicity:} We used the HOT dataset \cite{hotdataset} consisting of 3480 social media comments annotated as toxic or not (i.e., \textbf{2 labels}) by crowdworkers. 

\textbf{Sentiment analysis:} We used the SST5 dataset \cite{socher2013recursive} for fine-grained sentiment analysis where each sentiment is assigned to one of the \textbf{5 labels}: very negative, somewhat negative, neutral, somewhat positive, or positive. We used the test set consisting of 2210 text movie reviews annotated by crowdworkers. 

\textbf{Rumor stance detection:} We used the RumorEval dataset \cite{gorrell2019semeval} containing 1675 tweet pairs where the relationship between tweets is annotated as support, query, comment, or deny \textbf{(4 labels)} by crowdworkers. The task is more complex than identifying the relationship between a tweet and a fixed target (e.g., Hillary Clinton), on which LLMs have achieved close to SOTA performance \cite{zhang2022would}.

\textbf{News frame identification:} We used the GVFC dataset \cite{liu2019detecting} consisting of 1301 news headlines where expert scholars labeled the framing of the news article into one of the \textbf{9 frame classes}, such as gun rights, public opinion, etc.

\begin{table}[]
\resizebox{\columnwidth}{!}{
\begin{tabular}{@{}lcccc@{}}
\toprule
\multicolumn{1}{c}{\textbf{Dataset (\#labels)}} & \textbf{\#instances} & \textbf{\#prompts}   & \textbf{\#LLMs}      & \textbf{\#annotations} \\ \midrule
Toxicity (2)                             & 3,480                & 16                   & 3                    & 167,040                \\
Sentiment (5)                           & 2,210                & 8                    & 3                    & 53,040                 \\
Rumour Stance (4)                       & 1,675                & 16                   & 3                    & 80,400                 \\
News frame (9)                           & 1,301                & 16                   & 3                    & 62,448                 \\ \midrule
\textbf{Total data}                  & \multicolumn{1}{l}{} & \multicolumn{1}{l}{} & \multicolumn{1}{l}{} & \textbf{362,928}       \\ \bottomrule
\end{tabular}
}
\caption{Summary of annotations generated during the experiment}
\label{tab:dataset}

\end{table}

Table \ref{tab:dataset} shows the complete statistics of our dataset. We follow a complete factorial design between different prompt designs (2*2*2*2), LLMs (3), and datasets (4), resulting in a \textbf{total of 362,928 annotations}. The class distribution for each dataset is also provided in Appendix Table \ref{tab:class-dist}. 

\subsection{Evaluation} 

\textbf{Parsing LLM output:} We used simple string matching to extract potential labels from the LLM's raw output by matching against the list of labels provided in the input. If the LLM was prompted to provide numerical scores, we also extracted any floats between $[0, 1]$ from the output. The floats were matched to their corresponding labels based on the order in which they appeared. We also removed any text after ``explanation'' if the model was prompted to explain its output.

\textbf{Measuring compliance:} When prompting for an output label, the LLM's output was considered compliant if a unique label matching the input labels was extracted from the output. For instance, a model compliant with the toxicity task returned "yes" or "no" labels. When prompting for numerical scores, the output was considered compliant if at least one valid label was extracted from the output and the sum of scores assigned to the extracted labels belonged to $[0.99, 1.01]$. We report our results by computing percentage compliance on the complete dataset.  

\textbf{Measuring accuracy:} To calculate accuracy, we compared the LLM-generated labels with the human annotations provided for each dataset. We reported both F1 score (macro) and percentage accuracy in our results. We included only compliant outputs when measuring accuracy.

\section{Results}
First, we compare the overall accuracy and compliance for ChatGPT, Falcon7b, and PaLM2 on each task, and then present a breakdown of their compliance and accuracy for different prompt designs.

% \begin{figure*}
%     \centering
%     \includegraphics[width=\textwidth]{figures/llm-grid.png}
%     \caption{Comparing LLMs: a) percentage compliance; b) accuracy and F1 score; c) examples of noncompliance}
%     \label{fig:llm-grid}
% \end{figure*}

\subsection{Comparing LLMs}

\begin{figure}
    \centering
    \includegraphics[width=0.7\columnwidth]{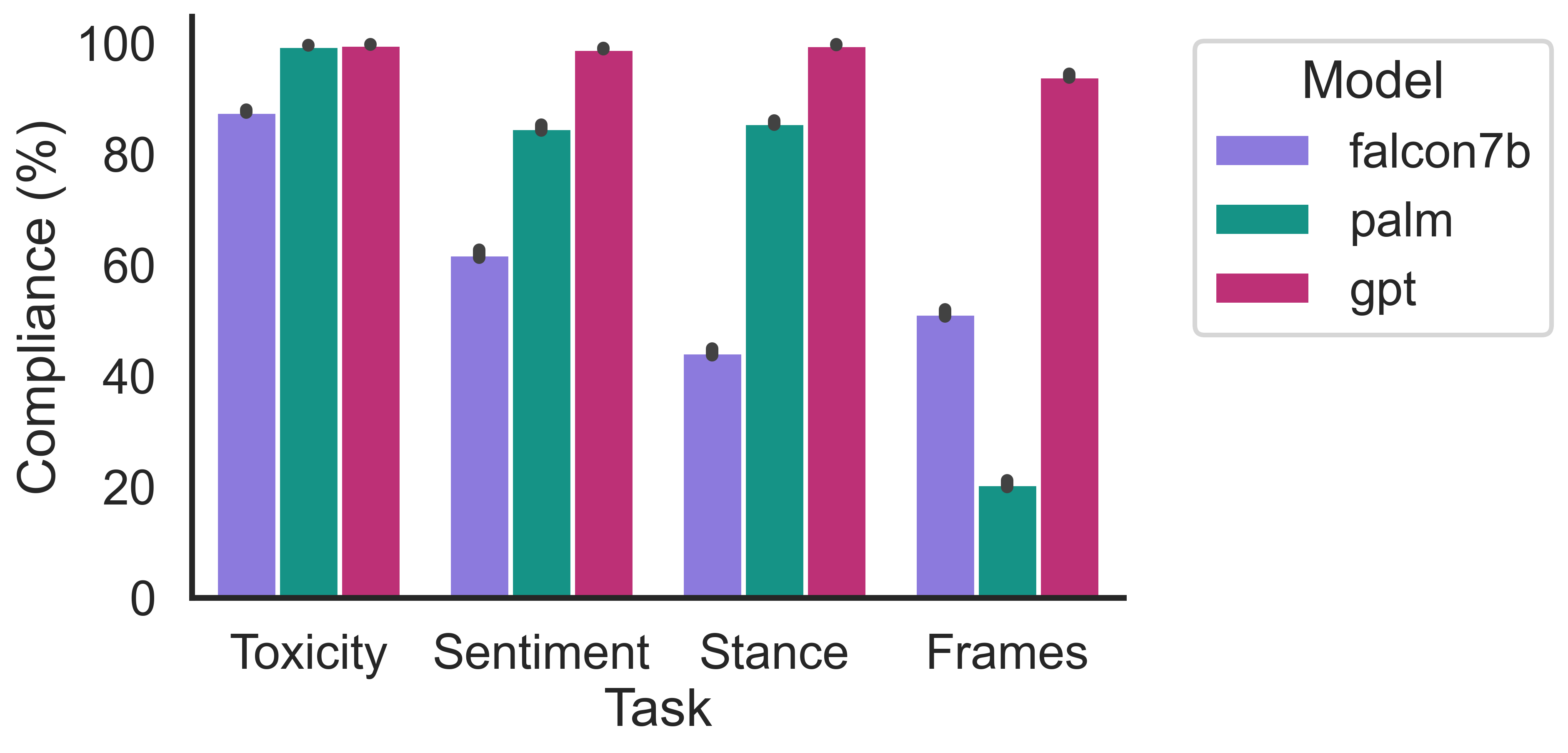}
    \caption{Percentage compliance for different tasks and LLMs.}
    \label{fig:model-compliance}
\end{figure}

% \usepackage{booktabs}
\begin{table}[]
\resizebox{\columnwidth}{!}{
\begin{tabular}{@{}lcccccccc@{}}
\toprule
\textbf{}                              & \multicolumn{2}{c}{\textbf{Toxicity}}           & \multicolumn{2}{c}{\textbf{Sentiment}}          & \multicolumn{2}{c}{\textbf{Stance}}             & \multicolumn{2}{c}{\textbf{Frames}} \\ \midrule
\multicolumn{1}{l|}{\textbf{}}         & \textbf{Acc} & \multicolumn{1}{c|}{\textbf{F1}} & \textbf{Acc} & \multicolumn{1}{c|}{\textbf{F1}} & \textbf{Acc} & \multicolumn{1}{c|}{\textbf{F1}} & \textbf{Acc}      & \textbf{F1}     \\ \midrule
\multicolumn{1}{l|}{\textbf{Falcon7b}} & 0.28         & \multicolumn{1}{c|}{0.28}        & 0.29         & \multicolumn{1}{c|}{0.29}        & 0.07         & \multicolumn{1}{c|}{0.07}        & 0.38              & 0.23            \\
\multicolumn{1}{l|}{\textbf{ChatGPT}}  & 0.71         & \multicolumn{1}{c|}{0.65}        & 0.41         & \multicolumn{1}{c|}{0.40}        & \textbf{0.62}         & \multicolumn{1}{c|}{0.38}        & 0.61              & 0.51            \\
\multicolumn{1}{c|}{\textbf{PaLM2}}    & \textbf{0.82}         & \multicolumn{1}{c|}{\textbf{0.73}}        & \textbf{0.53}         & \multicolumn{1}{c|}{\textbf{0.48}}        & 0.61         & \multicolumn{1}{c|}{\textbf{0.44}}        & \textbf{0.62}              & \textbf{0.54}            \\ \bottomrule
\end{tabular}
}
\caption{Percentage accuracy and F1 (macro) score for different tasks and LLMs}
\label{tab:models-acc}
\end{table}

\subsubsection{Compliance} 
\label{sec:llm-compliance}
\begin{figure*}
    \centering
    \includegraphics[width=0.8\textwidth]{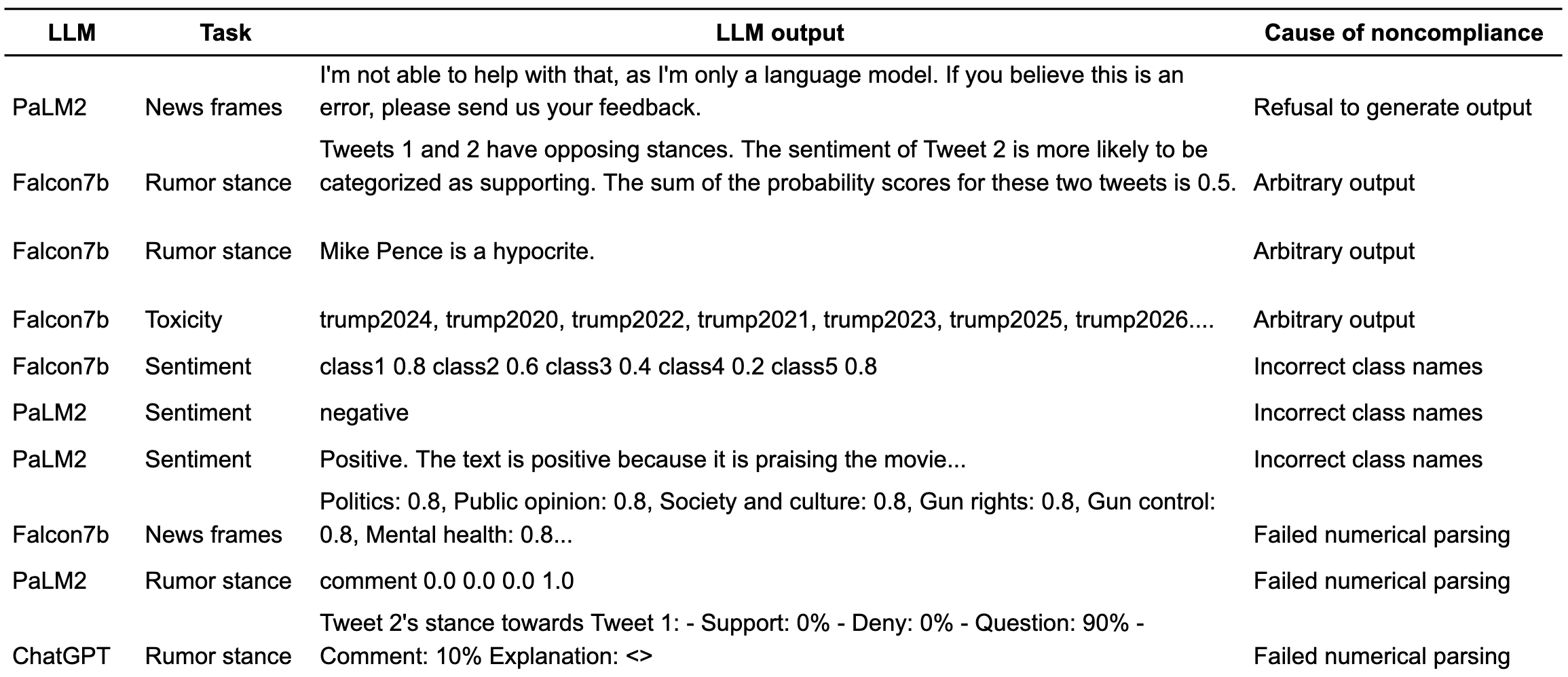}
    \caption{Examples demonstrating LLM noncompliance}
    \label{fig:noncomp-examples}
\end{figure*}

 We find that ChatGPT is the most compliant model for all tasks (see Figure \ref{fig:model-compliance}). The smallest model, Falcon7b, is the least compliant on all tasks except news frame identification, for which PaLM2 is the least compliant due to frequent refusals. Figure \ref{fig:noncomp-examples} shows PaLM2's refusal to perform news frame identification and other examples of noncompliance. Falcon7b often ignored instructions and generated arbitrary outputs. In other cases, LLMs generated incorrect class labels or failed to comply with numerical rules. 
 
 % LH: redundant with methods: Other causes of noncompliance include LLMs generating incorrect class labels or failing to comply with numerical rules. 
 
 % it refused to perform the task and responded with \emph{``I'm not able to help with that, as I'm only a language model. If you believe this is an error, please send us your feedback.''}. Other causes of noncompliance include LLMs generating incorrect class labels and LLMs failing to comply with numerical rules (see Figure \ref{fig:noncomp-examples}). 

\subsubsection{Accuracy} 
Table \ref{tab:models-acc} shows the overall accuracy and F1 score (macro) for all LLMs and tasks. Comparing both accuracy and F1 score, PaLM2 is the best-performing model on all tasks. Falcon7b is significantly less accurate than ChatGPT and PaLM2. 

\subsection{Comparing Prompts}

\subsubsection{Compliance}

\begin{table*}[]
\centering
\resizebox{\textwidth}{!}{
\begin{tabular}{@{}lccc|ccc|ccc|ccc@{}}
\toprule
\multicolumn{1}{c}{} & \multicolumn{3}{c|}{\textbf{Toxicity}}                & \multicolumn{3}{c|}{\textbf{Sentiment}}               & \multicolumn{3}{c|}{\textbf{Rumor stance}}            & \multicolumn{3}{c}{\textbf{News Frames}}              \\ \midrule
\multicolumn{1}{c}{\textbf{}}              & \textbf{Falcon7b} & \textbf{PaLM2} & \textbf{ChatGPT} & \textbf{Falcon7b} & \textbf{PaLM2} & \textbf{ChatGPT} & \textbf{Falcon7b} & \textbf{PaLM2} & \textbf{ChatGPT} & \textbf{Falcon7b} & \textbf{PaLM2} & \textbf{ChatGPT} \\ \midrule
Definition (yes)                           & 87.77             & 99.68          & 99.71            & ---              & ---           & ---             & 42.73             & 80.63          & 99.68            & 37.74             & 0.73           & \textbf{98.53}            \\
Definition (no)                            & 87.68             & 99.58          & 99.91            & 62.05             & 84.78          & 99.06            & \textbf{45.93}             & \textbf{90.77}          & 99.90            & \textbf{64.98}             & \textbf{40.43}          & 89.78            \\ \midrule
Explanation (yes)                          & 87.07             & 99.55          & 99.71            & \textbf{64.80}             & \textbf{91.65}          & 98.17            & \textbf{54.90}             & 86.29          & 99.62            & \textbf{63.90}             & 21.14          & \textbf{95.80}            \\
Explanation (no)                           & 88.38             & 99.72          & 99.91            & 59.31             & 77.91          & 99.95            & 33.76             & 85.10          & 99.96            & 38.83             & 20.02          & 92.51            \\ \midrule
Output Type (label)                        & 87.09             & 99.57          & 99.89            & \textbf{83.22}             & 77.01          & 100.00           & \textbf{71.85}             & \textbf{97.37}          & 99.98            & \textbf{83.29}             & 19.05          & \textbf{99.93}            \\
Output Type (score)                        & 88.35             & 99.70          & 99.73            & 40.88             & \textbf{92.55}          & 98.12            & 16.81             & 74.02          & 99.60            & 19.43             & 22.11          & 88.37            \\ \midrule
Length (standard)                          & 88.11             & 99.58          & 99.99            & \textbf{63.52}             & \textbf{95.97}          & 99.85            & 42.20             & 74.43          & 99.94            & 50.73             & \textbf{26.30}          & 93.51            \\
Length (concise)                           & 87.34             & 99.69          & 99.63            & 60.59             & 73.59          & 98.27            & \textbf{46.46}             & \textbf{96.96}          & 99.63            & 52.00             & 14.86          & 94.79            \\ \bottomrule
\end{tabular}
}
\caption{LLM percentage compliance for different prompt designs The more compliant variant of a prompt feature is highlighted in bold ($\Delta$ > 2\%)}
\label{tab:prompt-comp}
\end{table*}

\textbf{Overview:} Table \ref{tab:prompt-comp} presents a breakdown of LLMs' compliance for different prompt designs. For toxicity annotations, a task with only two class labels, compliance is high for all prompt and model combinations. But for other tasks, compliance varies by prompt and model. We also find that models exhibit differing levels of compliance for different prompts, with ChatGPT being the most consistent across prompts. Below, we highlight a few key observations for individual prompt designs.

\textbf{Definition (yes/no):} Prompting with class definitions reduces compliance for both Falcon7b and PaLM2 on rumor stance (4 classes) and news frame annotations (9 classes). Multiple class definitions make input instructions more complex and difficult to follow. We do not observe this relationship between definitions and compliance for ChatGPT which has demonstrated superior reasoning and dialogue capability compared to other LLMs \citep{qin-etal-2023-chatgpt}.

\begin{figure}
    \centering
\includegraphics[width=0.8\columnwidth]{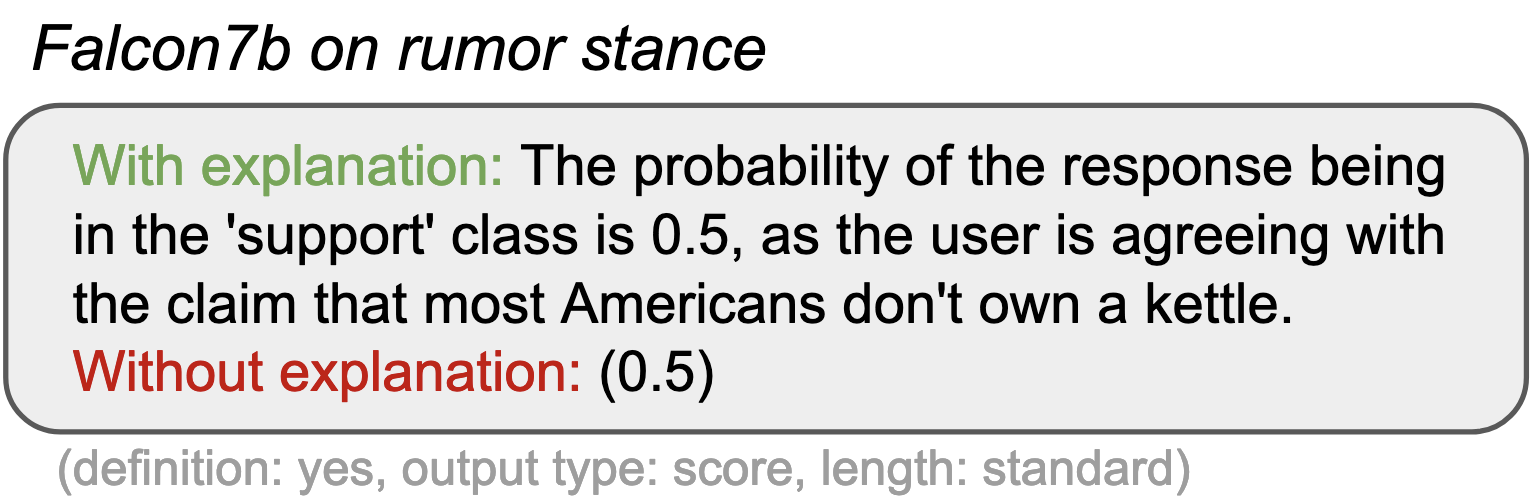}
    \caption{Falcon7b's response on the same data when prompted with/without explanation}
    \label{fig:explanation-ex}
\end{figure}

\textbf{Explanation (yes/no):} Prompting for an explanation in the output increases compliance for some task and model combinations: PaLM2 (sentiment), ChatGPT (news frames), and Falcon7b (sentiment, rumor stance, and news frames). Specific examples (see Figure \ref{fig:explanation-ex}) suggest that LLMs are less likely to respond with nonexistent or missing class labels when prompted to explain their output. 

\textbf{Prompt length (standard/concise):} Prompting ChatGPT with concise prompts has little impact on its compliance. This offers a significant cost advantage as ChatGPT's costs depend on the number of input tokens, and the concise version of a prompt on average contains 40\% fewer tokens. We do not observe this relationship between prompt length and compliance for other LLMs. 

\textbf{Output type (label/score):} Prompting for numerical scores instead of label class decreases compliance for Falcon7b (sentiment and rumor stance), PaLM2 (rumor stance), and ChatGPT (news frame). Noncompliance is often due to LLMs assigning the same score to each label, or providing scores with sum greater than 1 (Figure \ref{fig:noncomp-examples}). This is expected given LLMs' limitations in understanding numerical rules \cite{zhao2023docmath}. 

While prompting PaLM2 for numerical scores increases compliance for sentiment annotations, examples show that PaLM2 sometimes responds with coarse sentiment labels (instead of fine-grained) leading to noncompliance. This, however, is less likely to happen when PaLM2 is prompted for numerical scores (detailed example provided in Appendix Table \ref{tab:palm_sentiment}).


\begin{table*}[]
\centering
\resizebox{\textwidth}{!}{
\begin{tabular}{@{}lccc|ccc|ccc|ccc@{}}
\toprule
\multicolumn{1}{c}{\textbf{}} & \multicolumn{3}{c|}{\textbf{Toxicity}}                & \multicolumn{3}{c|}{\textbf{Sentiment}}               & \multicolumn{3}{c|}{\textbf{Rumor stance}}            & \multicolumn{3}{c}{\textbf{Frames}}                   \\ \midrule
\multicolumn{1}{c}{\textbf{}} & \textbf{Falcon7b} & \textbf{PaLM2} & \textbf{ChatGPT} & \textbf{Falcon7b} & \textbf{PaLM2} & \textbf{ChatGPT} & \textbf{Falcon7b} & \textbf{PaLM2} & \textbf{ChatGPT} & \textbf{Falcon7b} & \textbf{PaLM2} & \textbf{ChatGPT} \\ \midrule
Definition (yes)              & 24.20             & 81.13          & \textbf{73.11}            & ---              & ---           & ---             & 7.27              & 60.87          & 61.51            & 38.06             & \textbf{76.32}          & \textbf{67.77}            \\
Definition (no)               & \textbf{31.36}             & 82.70          & 69.45            & 28.78             & 53.23          & 41.20            & 7.35              & 61.65          & 61.83            & 37.71             & 62.05          & 53.18            \\ \midrule
Explanation (yes)             & 23.23             & 81.13          & 71.22            & \textbf{34.11}             & \textbf{55.70}          & 36.21            & 7.28              & 60.04          & \textbf{63.72}            & \textbf{41.86}             & \textbf{68.00}          & 54.88            \\
Explanation (no)              & \textbf{32.25}             & 82.70          & 71.34            & 22.96             & 50.33          & \textbf{46.11}            & 7.37              & \textbf{62.55}          & 59.62            & 31.21             & 56.29          & \textbf{66.96}            \\ \midrule
Output type (label)           & 17.68             & \textbf{85.09}          & \textbf{73.99}            & \textbf{31.00}             & \textbf{57.52}          & \textbf{46.39}            & 7.25              & \textbf{67.44}          & \textbf{69.87}            & 37.70             & \textbf{63.64}          & \textbf{67.49}            \\
output type (score)           & \textbf{37.73}             & 78.75          & 68.56            & 24.27             & 49.66          & 35.91            & 7.55              & 53.18          & 53.44            & 38.43             & 61.15          & 53.26            \\ \midrule
Length (standard)             & \textbf{31.85}             & \textbf{85.35}          & \textbf{73.38}            & 27.12             & 51.45          & 39.46            & 7.57              & 61.62          & \textbf{63.38}            & 33.38             & 61.64          & \textbf{63.57}            \\
Length (concise)              & 23.67             & 78.48          & 69.17            & \textbf{30.53}             & \textbf{55.56}          & \textbf{42.97}            & 7.07              & 61.03          & 59.95            & \textbf{42.18}             & 63.48          & 58.10            \\ \bottomrule
\end{tabular}
}
\caption{LLM percentage accuracy for different prompt designs. The more accurate variant of a prompt feature is highlighted in bold ($\Delta$ > 2\%)}
\label{tab:prompt-acc}
\end{table*}

\begin{table}[]
  \resizebox{\columnwidth}{!}{
    \begin{tabular}{@{}clccc@{}}
\toprule
\textbf{}                                                                                 & \multicolumn{1}{c}{\textbf{Label}} & \textbf{Explanation (yes)} & \textbf{Explanation (no)} & \textbf{$\Delta$}                \\ \midrule
                                                                                          & very positive                      & 1.43                       & 5.05                      & {\color[HTML]{EA4335} -3.62}  \\
                                                                                          & somewhat positive                  & 16.74                      & 33.25                     & {\color[HTML]{EA4335} -16.51} \\
                                                                                          & neutral                            & 54.37                      & 19.68                     & {\color[HTML]{34A853} 34.69}  \\
                                                                                          & somewhat negative                  & 19.57                      & 26.12                     & {\color[HTML]{EA4335} -6.55}  \\
\multirow{-5}{*}{\textbf{\begin{tabular}[c]{@{}c@{}}ChatGPT on\\ Sentiment\end{tabular}}} & very negative                      & 7.89                       & 15.90                     & {\color[HTML]{EA4335} -8.01}  \\ \midrule
                                                                                          & True                               & 91.59                      & 78.42                     & {\color[HTML]{34A853} 13.17}  \\
\multirow{-2}{*}{\textbf{\begin{tabular}[c]{@{}c@{}}Falcon7b on\\ Toxicity\end{tabular}}} & False                              & 8.41                       & 21.58                     & {\color[HTML]{EA4335} -13.17} \\ \bottomrule
\end{tabular}
        }
    \caption{Percentage distribution of generated labels when prompting LLMs with or without explanations}
    \label{tab:label-shift}
\end{table}

\subsubsection{Accuracy}

% \textbf{Overview:} 
Table \ref{tab:prompt-acc} presents a breakdown of LLMs' accuracy (F1 scores provided in the Appendix Table \ref{tab:prompt-f1}) for different prompt designs. We find that the impact of prompt design on accuracy is highly model and task dependent. Below, we highlight a few key observations for individual prompt designs.  

\textbf{Definition (yes/no):} Prompting with class definitions increases accuracy for ChatGPT (toxicity and news frames) and PaLM2 (news frames)\footnote{Although PaLM2's accuracy is measured on a very small subset of the complete data (<1\%) on which the model is compliant}. What is considered ``toxic'' can vary widely \cite{hotdataset}, and news frames can be defined in multiple ways \cite{nicholls2021computational}. Providing definitions can standardize these interpretations, leading to more accurate outputs from LLMs. We do not observe this trend for the smaller model, Falcon7b.

\textbf{Prompt length (standard/concise):} Prompting with concise prompts results in sentiment annotations with the same or higher accuracy for all LLMs. This is advantageous as concise prompts can reduce the costs of annotation. However, for toxicity annotations, concise prompts lead to lower accuracy for all LLMs, highlighting a tradeoff between cost and quality. While concise prompts can be efficient and cost-saving for some tasks, more detailed prompts may be necessary for achieving higher accuracy on complex tasks, such as toxicity.

\textbf{Output type (label/score):} Prompting for numerical scores instead of label class decreases the accuracy for all LLMs on all tasks (except for Falcon7b on toxicity). 

\textbf{Explanation (yes/no):} Prompting for an explanation in the output has a mixed impact on the accuracy of annotations depending on tasks and LLMs. In particular, prompting ChatGPT for explanation reduces the accuracy of sentiment and news frames annotations. Prompting Falcon7b for explanation also reduces accuracy of toxicity annotations. This undesirable impact undermines the potential of LLMs at generating human-like explanations \cite{huang2023chatgpt}. 

Further investigation shows that the impact of prompting with explanations on accuracy can be attributed to a major change in the distribution of LLM-generated labels. Table \ref{tab:label-shift} shows two examples. For sentiment labels generated by ChatGPT and toxicity labels generated by Falcon7b, the class distributions differed significantly depending on whether the model was prompted to provide an explanation or not.

\section{Discussion and Conclusion}

We empirically analyze the impact of prompt design on the quality of LLM-generated annotations using multiple LLMs and a diverse set of tasks. Our analysis reveals evidence-driven best practices for designing effective prompts. For example, prompting LLMs to explain their output improves their compliance with prompt instructions. Our findings also uncover inconsistencies in the impact of prompt design. For instance, prompting with the definition of toxicity improved the accuracy for ChatGPT, but not for PaLM2 or Falcon7b. Such inconsistencies highlight the need for researchers to carefully consider their prompt choices, as arbitrary decisions can influence conclusions about LLM performance, particularly when comparing different models.

Additionally, different prompt designs can also cause large shifts in the distributions of LLM-generated annotations, which can substantively affect social science research results. In the sections that follow, we unpack some of these implications and highlight potential directions for future research. We hope our study will serve as a foundation for further exploration into developing more effective and nuanced prompts for utilizing LLMs across various domains.

\subsection{Context-dependent Prompt Design}

In many cases, researchers or practitioners design LLM prompts that are highly influenced by their requirements and the context. For instance, \citet{li2024hot} prompted ChatGPT to provide probabilities of content being harmful or toxic instead of directly labeling the content. They argue that content moderators can use these numerical scores to better control content filtering. However, our findings indicate that prompting LLMs for numerical scores almost always leads to lower compliance across all models and tasks. This is not surprising given LLMs' limited numerical reasoning capabilities \cite{zhao2023docmath}. Nonetheless, it is crucial to recognize this limitation as addressing non-compliance will require additional resources.

In scenarios where output labels can be interpreted in multiple ways, such as news frames \cite{nicholls2021computational} or toxicity \cite{hotdataset}, practitioners might benefit from providing their own definitions to generate more accurate annotations. When accountability is a priority, practitioners may also want LLMs to explain their output. Interestingly, prompting for explanations can lead to more compliant outputs since LLMs are more likely to mention the correct class labels. However, as noted above, this can also cause significant shifts in the distribution of generated labels, which we discuss further below. 

\subsection{Implications of Prompting for CSS}
\label{sec:css-implications}

The success of LLM prompting has led to many applications in social science research \cite{ziems2024can}, such as monitoring public opinion \cite{zhang2022would} and quantifying online toxic content \cite{li2024hot}. Our results indicate that conclusions drawn from such research are highly dependent on the prompt design. For example, when annotating sentiment labels without prompting for an explanation, ChatGPT annotated $\sim$19\% of the data as neutral. However, when prompted to explain its output, ChatGPT labeled over 54\% of the data as neutral (see Table \ref{tab:label-shift}). Given the widespread applications of sentiment analysis for monitoring public opinion, such large systemic shifts in reported sentiment labels can lead to significant over- or underrepresentation of opinions. Furthermore, models trained using LLM-generated datasets will likely perpetuate these shifts in distribution. For instance, Falcon7b annotated 13\% more content as toxic when prompted to explain its output. Training a content moderation model on this dataset, or using Falcon7b directly, could result in more content being filtered or removed, exacerbating concerns of overmoderation \cite{ferrara2023should}. 

The reasons behind these shifts are unclear and could be due to differences in model architecture or the nature of training, including reinforcement learning from human feedback (RLHF). Nevertheless, social scientists should be cautious about the downstream impact of prompt variations on understanding social phenomena. When using LLMs, rather than using a single prompt unquestioningly, researchers should carefully evaluate several prompting strategies within their domain of interest, and potentially use several prompts for robustness when making claims. 

Future work should also explore potential strategies for mitigating these issues. One potential approach is to combine results across several prompts, similar to how crowdsourcing involves independent raters annotating the same content to improve the quality of annotations. 
Recent research \cite{echterhoff2024cognitive} shows that LLMs are capable of self-help debiasing to mitigate cognitive biases in input prompts. Therefore, exploring whether models can identify and self-correct large shifts in the distribution of generated labels induced by prompt variations could be a promising direction.

% FROM WOAH POSTER
% We find that the choice of LLM is more strongly associated with the annotation quality than prompt design. Within prompt design, cost saving concise prompts can provide compliant outputs although with lower accuracy. Compliance and accuracy offer different tradeoffs for practitioners, compliance can be identified without human annotations but addressing noncompliance may require chain of thought prompting \cite{wei2022chain}. Accuracy cannot to measured without human annotations but somewhat inaccurate data can still be used for training new models. This work is part of an ongoing project, and we are conducting a detailed cost vs. quality analysis to inform LLM-based data annotation tasks. Furthermore, we plan to investigate how model choice and prompt design affect downstream analyses of toxic and abusive language online. 

\section{Limitations}

Conducting rigorous evaluation of LLMs is challenging because we cannot determine whether these models have been exposed to our chosen datasets during their training phases, particularly popular datasets like SST-5. However, the fact that LLMs in our study fail to surpass existing baselines and show performance variability with different prompt designs suggests that any potential data leakage had limited impact on our findings.

Due to limited API availability and compute resources, we restricted our study to three LLMs. We only considered LLMs that were released as of June 2023. Due to our large-scale multifactorial design (2x2x2x2), we excluded more expensive models, such as Claude or GPT-4. Given our findings that the impact of prompt design depends on individual models, including their architecture and training data, we caution readers from generalizing our findings to other LLMs. Nevertheless, our inclusion of a smaller model, Falcon7b, reveals that its compliance decreases drastically as input prompts become more complex, such as when prompting for numerical scores. This finding underscores the need for future research to investigate alternative prompting techniques better suited for smaller, more affordable models.

Our study, with 16 prompts, is an extensive comparison of LLMs. However, the space of prompting is vast, and many variations for each prompt aspect are possible. Our results analyzed one prompt variation at a time, leaving open the possibility that different prompt variations may interact and produce different impacts. 

Despite these limitations, our research provides a foundation for future studies to explore additional prompt designs and investigate the interactions between different design choices. We hope our findings will inspire continued research in this area, leading to more effective and nuanced prompting strategies for a variety of LLMs.

\bibliography{anthology, custom}

\appendix

\section{Appendix}
\label{sec:appendix}

\begin{table*}[]
\centering
\caption{Class distribution in original datasets used for our experiment}
\label{tab:class-dist}

\begin{tabular}{@{}lclll@{}}
\toprule
\multicolumn{1}{c}{\textbf{Dataset}} & \textbf{\#instances} & \multicolumn{1}{c}{\textbf{\#prompts}} & \multicolumn{1}{c}{\textbf{\#LLMs}} & \multicolumn{1}{c}{\textbf{\#annotations}} \\ \midrule
\textbf{Toxicity}                    & \textbf{3480}        & \multicolumn{1}{c}{\textbf{16}}        & \multicolumn{1}{c}{\textbf{3}}      & \multicolumn{1}{c}{\textbf{167,040}}       \\
no                                   & 2885                 &                                        &                                     &                                            \\
yes                                  & 595                  &                                        &                                     &                                            \\
\textbf{Sentiment}                   & \textbf{2210}        & \multicolumn{1}{c}{\textbf{8}}         & \multicolumn{1}{c}{\textbf{3}}      & \multicolumn{1}{c}{\textbf{53,040}}        \\
very positive                        & 399                  &                                        &                                     &                                            \\
somewhat positive                    & 510                  &                                        &                                     &                                            \\
neutral                              & 389                  &                                        &                                     &                                            \\
somewhat negative                    & 633                  &                                        &                                     &                                            \\
very negative                        & 279                  &                                        &                                     &                                            \\
\textbf{Rumour Stance}               & \textbf{1675}        & \multicolumn{1}{c}{\textbf{16}}        & \multicolumn{1}{c}{\textbf{3}}      & \multicolumn{1}{c}{\textbf{80,400}}        \\
comment                              & 1405                 &                                        &                                     &                                            \\
support                              & 104                  &                                        &                                     &                                            \\
deny                                 & 100                  &                                        &                                     &                                            \\
query                                & 66                   &                                        &                                     &                                            \\
\textbf{News frame}                  & \textbf{1301}        & \multicolumn{1}{c}{\textbf{16}}        & \multicolumn{1}{c}{\textbf{3}}      & \multicolumn{1}{c}{\textbf{62,448}}        \\
Politics                             & 380                  &                                        &                                     &                                            \\
Public opinion                       & 237                  &                                        &                                     &                                            \\
Gun control                          & 215                  &                                        &                                     &                                            \\
School and public space              & 130                  &                                        &                                     &                                            \\
Race and ethnicity                   & 114                  &                                        &                                     &                                            \\
Economic consequences                & 81                   &                                        &                                     &                                            \\
Mental health                        & 65                   &                                        &                                     &                                            \\
Society and culture                  & 41                   &                                        &                                     &                                            \\
Gun rights                           & 38                   &                                        &                                     &                                            \\ \midrule
\textbf{Total}                       & \multicolumn{1}{l}{} &                                        &                                     & \multicolumn{1}{c}{\textbf{362,928}}       \\ \bottomrule
\end{tabular}
\end{table*}

% \usepackage{booktabs}
\begin{table*}[]
\caption{Example output generated by PaLM2 on sentiment annotation. PaLM2 is more likely to generate (compliant) fine-grained sentiment labels when prompted for numerical scores.}
\label{tab:palm_sentiment}
\resizebox{\textwidth}{!}{
\begin{tabular}{@{}cclc@{}}
\toprule
\multicolumn{1}{c}{\textbf{data\_id}} & \multicolumn{1}{c}{\textbf{output type}} & \multicolumn{1}{c}{\textbf{raw output}}                                                                                                                              & \multicolumn{1}{c}{\textbf{is\_compliant}} \\ \midrule
1820                                  & label                                    & negative                                                                                                                                                             & 0                                          \\
1820                                  & label                                    & negative                                                                                                                                                             & 0                                          \\
1820                                  & label                                    & negative, The text is negative because it is about a fish dying.                                                                                                     & 0                                          \\
1820                                  & label                                    & Negative. The text is about a fish out of water, which is a metaphor for someone who is not in their element and is therefore likely to fail.                        & 0                                          \\
1820                                  & score                                    & very negative 0.9999999999999999 very negative 0.9999999999999999 very negative 0.9999999999999999 very negative 0.9999999999999999 & 1                                          \\
1820                                  & score                                    & very negative 0.9999999999999999, somewhat negative 0.0, neutral 0.0, somewhat positive 0.0, very positive 0.0                                                       & 1                                          \\
1820                                  & score                                    & very negative 0.99; The text is very negative because it is about a fish that is out of water and will die.                                                          & 1                                          \\
1820                                  & score                                    & very negative 0.99 , The text is about a fish out of water. Fish out of water usually die. So the text is very negative.                                             & 1                                          \\ \bottomrule
\end{tabular}
}
\end{table*}

% \usepackage{booktabs}
\begin{table*}[]
\centering
\caption{LLM F1 scores for different prompt designs}
\label{tab:prompt-f1}
\resizebox{\textwidth}{!}{
\begin{tabular}{@{}lccc|ccc|ccc|ccc@{}}
\toprule
\multicolumn{1}{c}{\textbf{}} & \multicolumn{3}{c|}{\textbf{Toxicity}}                & \multicolumn{3}{c|}{\textbf{Sentiment}}               & \multicolumn{3}{c|}{\textbf{Rumor stance}}            & \multicolumn{3}{c}{\textbf{News frames}}              \\ \midrule
\multicolumn{1}{c}{\textbf{}} & \textbf{Falcon7b} & \textbf{PaLM2} & \textbf{ChatGPT} & \textbf{Falcon7b} & \textbf{PaLM2} & \textbf{ChatGPT} & \textbf{Falcon7b} & \textbf{PaLM2} & \textbf{ChatGPT} & \textbf{Falcon7b} & \textbf{PaLM2} & \textbf{ChatGPT} \\ \midrule
Definition (yes)              & 0.24              & 0.72           & 0.67             & ---               & ---            & ---              & 0.06              & 0.47           & 0.38             & 0.25              & 0.53           & 0.62             \\
Definition (no)               & 0.31              & 0.74           & 0.64             & 0.29              & 0.48           & 0.40             & 0.07              & 0.42           & 0.37             & 0.21              & 0.54           & 0.47             \\ \midrule
Explanation (yes)             & 0.23              & 0.72           & 0.65             & 0.34              & 0.51           & 0.34             & 0.07              & 0.45           & 0.39             & 0.26              & 0.60           & 0.48             \\
Explanation (no)              & 0.32              & 0.74           & 0.65             & 0.19              & 0.46           & 0.44             & 0.05              & 0.44           & 0.37             & 0.19              & 0.46           & 0.62             \\ \midrule
Output type (label)           & 0.16              & 0.76           & 0.67             & 0.31              & 0.53           & 0.42             & 0.06              & 0.50           & 0.42             & 0.22              & 0.56           & 0.58             \\
output type (score)           & 0.37              & 0.70           & 0.63             & 0.23              & 0.44           & 0.36             & 0.08              & 0.38           & 0.33             & 0.23              & 0.53           & 0.44             \\ \midrule
Length (standard)             & 0.32              & 0.76           & 0.67             & 0.24              & 0.46           & 0.38             & 0.07              & 0.44           & 0.38             & 0.21              & 0.52           & 0.52             \\
Length (concise)              & 0.23              & 0.70           & 0.64             & 0.30              & 0.51           & 0.41             & 0.07              & 0.45           & 0.37             & 0.24              & 0.60           & 0.51             \\ \bottomrule
\end{tabular}
}

\end{table*}

\subsection{Note on news frame identification task}

The input prompt for news frame identification instructed LLMs to provide up to two frame classes in their outputs. This prompt design is based on the instructions given to human annotators when the dataset was initially annotated in prior work. However, the original paper's authors noted that less than 10\% of the data was assigned more than one class. Consequently, the computational model introduced in the paper, and subsequent models trained on this data, treated the task as a single-class label and measured performance using only the first label provided by the human annotators. For consistency, we followed the same approach in our analysis and considered news frame identification as a single-label task.

\clearpage
\onecolumn
\begin{longtable}{|p{0.04\textwidth}|p{0.26\textwidth}|p{0.6\textwidth}|}

\caption{Set of prompts used in the experiment}
\label{tab:all-prompts}

\\ \hline
   & Prompt key                                             & Prompt string                                                                                                                                                                                                                                                                                                                                                                                                                                                                                                                                                                                                                                                                                                                                                                                                                                                                                                                                                                                                                                                                                                                                                                                                                                                                                                                                                                                                                                                                                                                                                                                                                                                                                                                                                                                                                                                                                                                                                                                                                                                                                                                                                                                                                                                                                                                                                                           \\ \hline
0  & toxicity-no\_definition-label-no\_explanation          & I want you to perform a data annotation task. Your task is to carefully read the text provided below and determine whether the text is toxic or not. In your output, I want you to respond with a yes if the text is toxic, otherwise respond with a no. Only respond with a yes or a no. Do not provide any other outputs or any explanation for your output. Text: \textless{}\textgreater{}                                                                                                                                                                                                                                                                                                                                                                                                                                                                                                                                                                                                                                                                                                                                                                                                                                                                                                                                                                                                                                                                                                                                                                                                                                                                                                                                                                                                                                                                                                                                                                                                                                                                                                                                                                                                                                                                                                                                                                                          \\ \hline
1  & toxicity-no\_definition-label-no\_explanation-concise  & Perform data annotation by determining if the text is toxic or not. Respond with only a yes or no. No other outputs or explanations needed. Text: \textless{}\textgreater{}                                                                                                                                                                                                                                                                                                                                                                                                                                                                                                                                                                                                                                                                                                                                                                                                                                                                                                                                                                                                                                                                                                                                                                                                                                                                                                                                                                                                                                                                                                                                                                                                                                                                                                                                                                                                                                                                                                                                                                                                                                                                                                                                                                                                             \\ \hline
2  & toxicity-no\_definition-label-explanation              & I want you to perform a data annotation task. Your task is to carefully read the text provided below and determine whether the text is toxic or not. In your output, I want you to respond with a yes if the text is toxic, otherwise respond with a no. Only respond with a yes or a no. In your output, I also want you to provide an explanation for the output. Provide your response in the first line and provide the explanation for your response in the second line. Text: \textless{}\textgreater{}                                                                                                                                                                                                                                                                                                                                                                                                                                                                                                                                                                                                                                                                                                                                                                                                                                                                                                                                                                                                                                                                                                                                                                                                                                                                                                                                                                                                                                                                                                                                                                                                                                                                                                                                                                                                                                                                           \\ \hline
3  & toxicity-no\_definition-label-explanation-concise      & Perform a data annotation task by determining if the provided text is toxic. Respond with a yes or no and provide an explanation for your response in two lines. Text: \textless{}\textgreater{}                                                                                                                                                                                                                                                                                                                                                                                                                                                                                                                                                                                                                                                                                                                                                                                                                                                                                                                                                                                                                                                                                                                                                                                                                                                                                                                                                                                                                                                                                                                                                                                                                                                                                                                                                                                                                                                                                                                                                                                                                                                                                                                                                                                        \\ \hline
4  & toxicity-no\_definition-score-no\_explanation          & I want you to perform a data annotation task. Your task is to carefully read the text provided below and determine whether the text is toxic or not. In your output, I want you to provide a number between 0 and 1 where the number represents the probability of the text being toxic. A higher score indicates a higher probability of the text being toxic. Only respond with a number between 0 and 1. Do not provide any other outputs or any explanation for your output. Text: \textless{}\textgreater{}                                                                                                                                                                                                                                                                                                                                                                                                                                                                                                                                                                                                                                                                                                                                                                                                                                                                                                                                                                                                                                                                                                                                                                                                                                                                                                                                                                                                                                                                                                                                                                                                                                                                                                                                                                                                                                                                        \\ \hline
5  & toxicity-no\_definition-score-no\_explanation-concise  & Perform data annotation by determining if the text is toxic. Provide a number between 0 and 1 indicating the probability of toxicity, with a higher score indicating higher probability. Only provide the number, no other outputs or explanations. Text: \textless{}\textgreater{}                                                                                                                                                                                                                                                                                                                                                                                                                                                                                                                                                                                                                                                                                                                                                                                                                                                                                                                                                                                                                                                                                                                                                                                                                                                                                                                                                                                                                                                                                                                                                                                                                                                                                                                                                                                                                                                                                                                                                                                                                                                                                                     \\ \hline
6  & toxicity-no\_definition-score-explanation              & I want you to perform a data annotation task. Your task is to carefully read the text provided below and determine whether the text is toxic or not. In your output, I want you to provide a number between 0 and 1 where the number represents the probability of the text being toxic. A higher score indicates a higher probability of the text being toxic. Only respond with a number between 0 and 1. In your output, I also want you to provide an explanation for the output. Provide your response in the first line and provide the explanation for your response in the second line. Text: \textless{}\textgreater{}                                                                                                                                                                                                                                                                                                                                                                                                                                                                                                                                                                                                                                                                                                                                                                                                                                                                                                                                                                                                                                                                                                                                                                                                                                                                                                                                                                                                                                                                                                                                                                                                                                                                                                                                                         \\ \hline
7  & toxicity-no\_definition-score-explanation-concise      & Perform data annotation by reading the text and determining its toxicity level with a number between 0 and 1. Higher scores indicate higher toxicity. Include an explanation for your response in the output. Text: \textless{}\textgreater{}                                                                                                                                                                                                                                                                                                                                                                                                                                                                                                                                                                                                                                                                                                                                                                                                                                                                                                                                                                                                                                                                                                                                                                                                                                                                                                                                                                                                                                                                                                                                                                                                                                                                                                                                                                                                                                                                                                                                                                                                                                                                                                                                           \\ \hline
8  & toxicity-definition-label-no\_explanation              & I want you to perform a data annotation task. Your task is to carefully read the text provided below and determine whether the text is toxic or not. For this task, I also want you to consider the following definition of toxicity: A rude, disrespectful, or unreasonable comment that is likely to make individuals leave a discussion. Annotate the text as toxic only if it satisfies the definition provided above, otherwise do not annotate it as toxic. In your output, I want you to respond with a yes if the text is toxic, otherwise respond with a no. Only respond with a yes or a no. Do not provide any other outputs or any explanation for your output. Text: \textless{}\textgreater{}                                                                                                                                                                                                                                                                                                                                                                                                                                                                                                                                                                                                                                                                                                                                                                                                                                                                                                                                                                                                                                                                                                                                                                                                                                                                                                                                                                                                                                                                                                                                                                                                                                                                             \\ \hline
9  & toxicity-definition-label-no\_explanation-concise      & Perform data annotation task. Read text and determine toxicity based on definition: rude, disrespectful, or unreasonable comment likely to make individuals leave discussion. Respond with yes if toxic, otherwise no. No other outputs or explanations. Text: \textless{}\textgreater{}                                                                                                                                                                                                                                                                                                                                                                                                                                                                                                                                                                                                                                                                                                                                                                                                                                                                                                                                                                                                                                                                                                                                                                                                                                                                                                                                                                                                                                                                                                                                                                                                                                                                                                                                                                                                                                                                                                                                                                                                                                                                                                \\ \hline
10 & toxicity-definition-label-explanation                  & I want you to perform a data annotation task. Your task is to carefully read the text provided below and determine whether the text is toxic or not. For this task, I also want you to consider the following definition of toxicity: A rude, disrespectful, or unreasonable comment that is likely to make individuals leave a discussion. Annotate the text as toxic only if it satisfies the definition provided above, otherwise do not annotate it as toxic. In your output, I want you to respond with a yes if the text is toxic, otherwise respond with a no. Only respond with a yes or a no. In your output, I also want you to provide an explanation for the output. Provide your response in the first line and provide the explanation for your response in the second line. Text: \textless{}\textgreater{}                                                                                                                                                                                                                                                                                                                                                                                                                                                                                                                                                                                                                                                                                                                                                                                                                                                                                                                                                                                                                                                                                                                                                                                                                                                                                                                                                                                                                                                                                                                                                              \\ \hline
11 & toxicity-definition-label-explanation-concise          & Perform a data annotation task by determining if the text below is toxic. Use the definition of toxicity as a rude, disrespectful, or unreasonable comment likely to make individuals leave a discussion. Respond with a yes or no and provide an explanation for your response. Text: \textless{}\textgreater{}                                                                                                                                                                                                                                                                                                                                                                                                                                                                                                                                                                                                                                                                                                                                                                                                                                                                                                                                                                                                                                                                                                                                                                                                                                                                                                                                                                                                                                                                                                                                                                                                                                                                                                                                                                                                                                                                                                                                                                                                                                                                        \\ \hline
12 & toxicity-definition-score-no\_explanation              & I want you to perform a data annotation task. Your task is to carefully read the text provided below and determine whether the text is toxic or not. For this task, I also want you to consider the following definition of toxicity: A rude, disrespectful, or unreasonable comment that is likely to make individuals leave a discussion. Annotate the text as toxic only if it satisfies the definition provided above, otherwise do not annotate it as toxic. In your output, I want you to provide a number between 0 and 1 where the number represents the probability of the text being toxic. A higher score indicates a higher probability of the text being toxic. Only respond with a number between 0 and 1. Do not provide any other outputs or any explanation for your output. Text: \textless{}\textgreater{}                                                                                                                                                                                                                                                                                                                                                                                                                                                                                                                                                                                                                                                                                                                                                                                                                                                                                                                                                                                                                                                                                                                                                                                                                                                                                                                                                                                                                                                                                                                                                           \\ \hline
13 & toxicity-definition-score-no\_explanation-concise      & Perform data annotation by determining if the text is toxic. Use the definition of toxicity as a rude, disrespectful, or unreasonable comment likely to make individuals leave a discussion. Annotate as toxic only if it meets this definition. Provide a number between 0 and 1 indicating the probability of toxicity, with a higher score indicating a higher probability. Do not provide any other outputs or explanations. Text: \textless{}\textgreater{}                                                                                                                                                                                                                                                                                                                                                                                                                                                                                                                                                                                                                                                                                                                                                                                                                                                                                                                                                                                                                                                                                                                                                                                                                                                                                                                                                                                                                                                                                                                                                                                                                                                                                                                                                                                                                                                                                                                        \\ \hline
14 & toxicity-definition-score-explanation                  & I want you to perform a data annotation task. Your task is to carefully read the text provided below and determine whether the text is toxic or not. For this task, I also want you to consider the following definition of toxicity: A rude, disrespectful, or unreasonable comment that is likely to make individuals leave a discussion. Annotate the text as toxic only if it satisfies the definition provided above, otherwise do not annotate it as toxic. In your output, I want you to provide a number between 0 and 1 where the number represents the probability of the text being toxic. A higher score indicates a higher probability of the text being toxic. Only respond with a number between 0 and 1. In your output, I also want you to provide an explanation for the output. Provide your response in the first line and provide the explanation for your response in the second line. Text: \textless{}\textgreater{}                                                                                                                                                                                                                                                                                                                                                                                                                                                                                                                                                                                                                                                                                                                                                                                                                                                                                                                                                                                                                                                                                                                                                                                                                                                                                                                                                                                                                                            \\ \hline
15 & toxicity-definition-score-explanation-concise          & Perform data annotation by determining if the text is toxic. Use the definition of toxicity as a rude, disrespectful, or unreasonable comment likely to make individuals leave a discussion. Annotate as toxic only if it meets this definition. Provide a probability score between 0 and 1, with a higher score indicating a higher probability of toxicity. Include an explanation for the score in the output. Text: \textless{}\textgreater{}                                                                                                                                                                                                                                                                                                                                                                                                                                                                                                                                                                                                                                                                                                                                                                                                                                                                                                                                                                                                                                                                                                                                                                                                                                                                                                                                                                                                                                                                                                                                                                                                                                                                                                                                                                                                                                                                                                                                      \\ \hline
16 & stance-no\_definition-label-no\_explanation            & I want you to perform a data annotation task. Your task is to carefully read two tweets and determine the stance of tweet 2 in response to tweet 1. Your response must belong to one of the four classes, depending on whether tweet 2 supports, denies, questions, or comments on tweet 1. In your output, only respond with the name of the class: support, deny, question, or comment, depending on the relation you identify between tweet 1 and tweet 2. Do not respond with any other output. Do not provide any other outputs or any explanation for your output. Tweet 1: \textless{}\textgreater Tweet 2: \textless{}\textgreater{}                                                                                                                                                                                                                                                                                                                                                                                                                                                                                                                                                                                                                                                                                                                                                                                                                                                                                                                                                                                                                                                                                                                                                                                                                                                                                                                                                                                                                                                                                                                                                                                                                                                                                                                                            \\ \hline
17 & stance-no\_definition-label-no\_explanation-concise    & Perform data annotation by reading two tweets and identifying the stance of tweet 2 towards tweet 1. Choose from four classes: support, deny, question, or comment. Only provide the name of the class in your output without any additional explanation. Tweet 1: \textless{}\textgreater Tweet 2: \textless{}\textgreater{}                                                                                                                                                                                                                                                                                                                                                                                                                                                                                                                                                                                                                                                                                                                                                                                                                                                                                                                                                                                                                                                                                                                                                                                                                                                                                                                                                                                                                                                                                                                                                                                                                                                                                                                                                                                                                                                                                                                                                                                                                                                           \\ \hline
18 & stance-no\_definition-label-explanation                & I want you to perform a data annotation task. Your task is to carefully read two tweets and determine the stance of tweet 2 in response to tweet 1. Your response must belong to one of the four classes, depending on whether tweet 2 supports, denies, questions, or comments on tweet 1. In your output, only respond with the name of the class: support, deny, question, or comment, depending on the relation you identify between tweet 1 and tweet 2. Do not respond with any other output. In your output, I also want you to provide an explanation for the output. Provide your response in the first line and provide the explanation for your response in the second line. Tweet 1: \textless{}\textgreater Tweet 2: \textless{}\textgreater{}                                                                                                                                                                                                                                                                                                                                                                                                                                                                                                                                                                                                                                                                                                                                                                                                                                                                                                                                                                                                                                                                                                                                                                                                                                                                                                                                                                                                                                                                                                                                                                                                                             \\ \hline
19 & stance-no\_definition-label-explanation-concise        & Perform a data annotation task by reading two tweets and identifying the stance of tweet 2 towards tweet 1. Choose from four classes: support, deny, question, or comment. Only provide the class name in your output and include an explanation in the second line. Tweet 1: \textless{}\textgreater Tweet 2: \textless{}\textgreater{}                                                                                                                                                                                                                                                                                                                                                                                                                                                                                                                                                                                                                                                                                                                                                                                                                                                                                                                                                                                                                                                                                                                                                                                                                                                                                                                                                                                                                                                                                                                                                                                                                                                                                                                                                                                                                                                                                                                                                                                                                                                \\ \hline
20 & stance-no\_definition-score-no\_explanation            & I want you to perform a data annotation task. Your task is to carefully read two tweets and determine the stance of tweet 2 in response to tweet 1. Your response must belong to one of the four classes, depending on whether tweet 2 supports, denies, questions, or comments on tweet 1. In your output, I want you to provide a probability score for each of the 4 classes. The probability of each class should be a number between 0 and 1 where higher numbers represent a higher probability of that class. Since there are only 4 possible classes, the sum of their probability scores should always be equal to 1. For each class, respond with the name of the class followed by its probability score in each line. Do not provide any other outputs or any explanation for your output. Tweet 1: \textless{}\textgreater Tweet 2: \textless{}\textgreater{}                                                                                                                                                                                                                                                                                                                                                                                                                                                                                                                                                                                                                                                                                                                                                                                                                                                                                                                                                                                                                                                                                                                                                                                                                                                                                                                                                                                                                                                                                                              \\ \hline
21 & stance-no\_definition-score-no\_explanation-concise    & Perform data annotation by reading 2 tweets and determining tweet 2's stance towards tweet 1. Choose from 4 classes: support, deny, question, or comment. Provide probability scores for each class, with higher numbers indicating higher probability. Total probability scores should equal 1. Output only class names and scores, no explanations. Tweet 1: \textless{}\textgreater Tweet 2: \textless{}\textgreater{}                                                                                                                                                                                                                                                                                                                                                                                                                                                                                                                                                                                                                                                                                                                                                                                                                                                                                                                                                                                                                                                                                                                                                                                                                                                                                                                                                                                                                                                                                                                                                                                                                                                                                                                                                                                                                                                                                                                                                               \\ \hline
22 & stance-no\_definition-score-explanation                & I want you to perform a data annotation task. Your task is to carefully read two tweets and determine the stance of tweet 2 in response to tweet 1. Your response must belong to one of the four classes, depending on whether tweet 2 supports, denies, questions, or comments on tweet 1. In your output, I want you to provide a probability score for each of the 4 classes. The probability of each class should be a number between 0 and 1 where higher numbers represent a higher probability of that class. Since there are only 4 possible classes, the sum of their probability scores should always be equal to 1. For each class, respond with the name of the class followed by its probability score in each line. In your output, I also want you to provide an explanation for the output. Provide your response in the first line and provide the explanation for your response in the second line. Tweet 1: \textless{}\textgreater Tweet 2: \textless{}\textgreater{}                                                                                                                                                                                                                                                                                                                                                                                                                                                                                                                                                                                                                                                                                                                                                                                                                                                                                                                                                                                                                                                                                                                                                                                                                                                                                                                                                                                               \\ \hline
23 & stance-no\_definition-score-explanation-concise        & Perform a data annotation task by analyzing two tweets and determining the stance of tweet 2 towards tweet 1. Categorize tweet 2 as supporting, denying, questioning, or commenting on tweet 1 and provide a probability score for each class. The sum of the probability scores should be 1. Output the name of the class and its probability score for each line. Additionally, provide an explanation for your response in the second line of the output. Tweet 1: \textless{}\textgreater Tweet 2: \textless{}\textgreater{}                                                                                                                                                                                                                                                                                                                                                                                                                                                                                                                                                                                                                                                                                                                                                                                                                                                                                                                                                                                                                                                                                                                                                                                                                                                                                                                                                                                                                                                                                                                                                                                                                                                                                                                                                                                                                                                        \\ \hline
24 & stance-definition-label-no\_explanation                & I want you to perform a data annotation task. Your task is to carefully read two tweets and determine the stance of tweet 2 in response to tweet 1. Your response must belong to one of the four classes, depending on whether tweet 2 supports, denies, questions, or comments on tweet 1. For this task, respond with support if the reply supports the claim, respond with deny if the reply disagrees with the claim, respond with question if the reply is asking for additional evidence in relation to the claim, and respond with comment if the reply is making its own claim without a clear contribution to assessing the veracity of the claim. You must follow the instructions mentioned above when providing your response. Do not provide a response that does not align with the instructions. In your output, only respond with the name of the class: support, deny, question, or comment, depending on the relation you identify between tweet 1 and tweet 2. Do not respond with any other output. Do not provide any other outputs or any explanation for your output. Tweet 1: \textless{}\textgreater Tweet 2: \textless{}\textgreater{}                                                                                                                                                                                                                                                                                                                                                                                                                                                                                                                                                                                                                                                                                                                                                                                                                                                                                                                                                                                                                                                                                                                                                                                                                        \\ \hline
25 & stance-definition-label-no\_explanation-concise        & Perform a data annotation task by reading two tweets and determining the stance of tweet 2 towards tweet 1. Choose from four classes: support, deny, question, or comment. Only respond with the name of the class that aligns with the instructions. Do not provide any other output or explanation. Tweet 1: \textless{}\textgreater Tweet 2: \textless{}\textgreater{}                                                                                                                                                                                                                                                                                                                                                                                                                                                                                                                                                                                                                                                                                                                                                                                                                                                                                                                                                                                                                                                                                                                                                                                                                                                                                                                                                                                                                                                                                                                                                                                                                                                                                                                                                                                                                                                                                                                                                                                                               \\ \hline
26 & stance-definition-label-explanation                    & I want you to perform a data annotation task. Your task is to carefully read two tweets and determine the stance of tweet 2 in response to tweet 1. Your response must belong to one of the four classes, depending on whether tweet 2 supports, denies, questions, or comments on tweet 1. For this task, respond with support if the reply supports the claim, respond with deny if the reply disagrees with the claim, respond with question if the reply is asking for additional evidence in relation to the claim, and respond with comment if the reply is making its own claim without a clear contribution to assessing the veracity of the claim. You must follow the instructions mentioned above when providing your response. Do not provide a response that does not align with the instructions. In your output, only respond with the name of the class: support, deny, question, or comment, depending on the relation you identify between tweet 1 and tweet 2. Do not respond with any other output. In your output, I also want you to provide an explanation for the output. Provide your response in the first line and provide the explanation for your response in the second line. Tweet 1: \textless{}\textgreater Tweet 2: \textless{}\textgreater{}                                                                                                                                                                                                                                                                                                                                                                                                                                                                                                                                                                                                                                                                                                                                                                                                                                                                                                                                                                                                                                                                                                         \\ \hline
27 & stance-definition-label-explanation-concise            & Perform data annotation by reading two tweets and identifying the stance of tweet 2 towards tweet 1. Choose from four classes: support, deny, question, or comment. Respond with only the class name and provide an explanation for your choice in the second line. Follow the instructions and do not deviate from them. Tweet 1: \textless{}\textgreater Tweet 2: \textless{}\textgreater{}                                                                                                                                                                                                                                                                                                                                                                                                                                                                                                                                                                                                                                                                                                                                                                                                                                                                                                                                                                                                                                                                                                                                                                                                                                                                                                                                                                                                                                                                                                                                                                                                                                                                                                                                                                                                                                                                                                                                                                                           \\ \hline
28 & stance-definition-score-no\_explanation                & I want you to perform a data annotation task. Your task is to carefully read two tweets and determine the stance of tweet 2 in response to tweet 1. Your response must belong to one of the four classes, depending on whether tweet 2 supports, denies, questions, or comments on tweet 1. For this task, respond with support if the reply supports the claim, respond with deny if the reply disagrees with the claim, respond with question if the reply is asking for additional evidence in relation to the claim, and respond with comment if the reply is making its own claim without a clear contribution to assessing the veracity of the claim. You must follow the instructions mentioned above when providing your response. Do not provide a response that does not align with the instructions. In your output, I want you to provide a probability score for each of the 4 classes. The probability of each class should be a number between 0 and 1 where higher numbers represent a higher probability of that class. Since there are only 4 possible classes, the sum of their probability scores should always be equal to 1. For each class, respond with the name of the class followed by its probability score in each line. Do not provide any other outputs or any explanation for your output. Tweet 1: \textless{}\textgreater Tweet 2: \textless{}\textgreater{}                                                                                                                                                                                                                                                                                                                                                                                                                                                                                                                                                                                                                                                                                                                                                                                                                                                                                                                                                                                          \\ \hline
29 & stance-definition-score-no\_explanation-concise        & Perform a data annotation task by reading two tweets and determining the stance of tweet 2 towards tweet 1. Choose from four classes: support, deny, question, or comment. Respond with the name of the class and a probability score between 0 and 1 for each class. The sum of the probability scores should be 1. Follow the instructions carefully and do not provide any additional outputs or explanations. Tweet 1: \textless{}\textgreater Tweet 2: \textless{}\textgreater{}                                                                                                                                                                                                                                                                                                                                                                                                                                                                                                                                                                                                                                                                                                                                                                                                                                                                                                                                                                                                                                                                                                                                                                                                                                                                                                                                                                                                                                                                                                                                                                                                                                                                                                                                                                                                                                                                                                   \\ \hline
30 & stance-definition-score-explanation                    & I want you to perform a data annotation task. Your task is to carefully read two tweets and determine the stance of tweet 2 in response to tweet 1. Your response must belong to one of the four classes, depending on whether tweet 2 supports, denies, questions, or comments on tweet 1. For this task, respond with support if the reply supports the claim, respond with deny if the reply disagrees with the claim, respond with question if the reply is asking for additional evidence in relation to the claim, and respond with comment if the reply is making its own claim without a clear contribution to assessing the veracity of the claim. You must follow the instructions mentioned above when providing your response. Do not provide a response that does not align with the instructions. In your output, I want you to provide a probability score for each of the 4 classes. The probability of each class should be a number between 0 and 1 where higher numbers represent a higher probability of that class. Since there are only 4 possible classes, the sum of their probability scores should always be equal to 1. For each class, respond with the name of the class followed by its probability score in each line. In your output, I also want you to provide an explanation for the output. Provide your response in the first line and provide the explanation for your response in the second line. Tweet 1: \textless{}\textgreater Tweet 2: \textless{}\textgreater{}                                                                                                                                                                                                                                                                                                                                                                                                                                                                                                                                                                                                                                                                                                                                                                                                                                                                           \\ \hline
31 & stance-definition-score-explanation-concise            & Perform data annotation by analyzing two tweets and determining the stance of tweet 2 towards tweet 1. Choose from four classes: support, deny, question, or comment. Respond with a probability score for each class, where higher numbers indicate higher probability. The sum of all probability scores should be 1. Follow the instructions carefully and provide an explanation for your response. Tweet 1: \textless{}\textgreater Tweet 2: \textless{}\textgreater{}                                                                                                                                                                                                                                                                                                                                                                                                                                                                                                                                                                                                                                                                                                                                                                                                                                                                                                                                                                                                                                                                                                                                                                                                                                                                                                                                                                                                                                                                                                                                                                                                                                                                                                                                                                                                                                                                                                             \\ \hline
32 & sentiment-no\_definition-label-no\_explanation         & I want you to perform a data annotation task. Your task is to carefully read the text and identify the polarity of the sentiment that is conveyed. Your response must belong to one of the five classes, depending on whether the text is very positive, somewhat positive, neutral, somewhat negative, or very negative. In your output, only respond with the name of the class: very positive, somewhat positive, neutral, somewhat negative, or very negative, depending on the sentiment that is conveyed in the text. Do not provide any other outputs or any explanation for your output. Text: \textless{}\textgreater{}                                                                                                                                                                                                                                                                                                                                                                                                                                                                                                                                                                                                                                                                                                                                                                                                                                                                                                                                                                                                                                                                                                                                                                                                                                                                                                                                                                                                                                                                                                                                                                                                                                                                                                                                                        \\ \hline
33 & sentiment-no\_definition-label-no\_explanation-concise & Perform data annotation by identifying sentiment polarity as very positive, somewhat positive, neutral, somewhat negative, or very negative. Only provide the name of the class in your output, without any additional explanation. Text: \textless{}\textgreater{}                                                                                                                                                                                                                                                                                                                                                                                                                                                                                                                                                                                                                                                                                                                                                                                                                                                                                                                                                                                                                                                                                                                                                                                                                                                                                                                                                                                                                                                                                                                                                                                                                                                                                                                                                                                                                                                                                                                                                                                                                                                                                                                     \\ \hline
34 & sentiment-no\_definition-label-explanation             & I want you to perform a data annotation task. Your task is to carefully read the text and identify the polarity of the sentiment that is conveyed. Your response must belong to one of the five classes, depending on whether the text is very positive, somewhat positive, neutral, somewhat negative, or very negative. In your output, only respond with the name of the class: very positive, somewhat positive, neutral, somewhat negative, or very negative, depending on the sentiment that is conveyed in the text. In your output, I also want you to provide an explanation for the output. Provide your response in the first line and provide the explanation for your response in the second line. Text: \textless{}\textgreater{}                                                                                                                                                                                                                                                                                                                                                                                                                                                                                                                                                                                                                                                                                                                                                                                                                                                                                                                                                                                                                                                                                                                                                                                                                                                                                                                                                                                                                                                                                                                                                                                                                                         \\ \hline
35 & sentiment-no\_definition-label-explanation-concise     & Perform a data annotation task by identifying the sentiment polarity of the text as very positive, somewhat positive, neutral, somewhat negative, or very negative. Provide only the class name and an explanation for your response in the first and second lines of your output, respectively. Text: \textless{}\textgreater{}                                                                                                                                                                                                                                                                                                                                                                                                                                                                                                                                                                                                                                                                                                                                                                                                                                                                                                                                                                                                                                                                                                                                                                                                                                                                                                                                                                                                                                                                                                                                                                                                                                                                                                                                                                                                                                                                                                                                                                                                                                                        \\ \hline
36 & sentiment-no\_definition-score-no\_explanation         & I want you to perform a data annotation task. Your task is to carefully read the text and identify the polarity of the sentiment that is conveyed. Your response must belong to one of the five classes, depending on whether the text is very positive, somewhat positive, neutral, somewhat negative, or very negative. In your output, I want you to provide a probability score for each of the 5 classes. The probability of each class should be a number between 0 and 1 where higher numbers represent a higher probability of that class. Since there are only 5 possible classes, the sum of their probability scores should always be equal to 1. For each class, respond with the name of the class followed by its probability score in each line. Do not provide any other outputs or any explanation for your output. Text: \textless{}\textgreater{}                                                                                                                                                                                                                                                                                                                                                                                                                                                                                                                                                                                                                                                                                                                                                                                                                                                                                                                                                                                                                                                                                                                                                                                                                                                                                                                                                                                                                                                                                                                    \\ \hline
37 & sentiment-no\_definition-score-no\_explanation-concise & Perform data annotation by identifying sentiment polarity as very positive, somewhat positive, neutral, somewhat negative, or very negative. Provide probability scores for each class, with higher numbers indicating higher probability. Sum of probability scores for all classes should be 1. Respond with class name and probability score for each line, without any additional output or explanation. Text: \textless{}\textgreater{}                                                                                                                                                                                                                                                                                                                                                                                                                                                                                                                                                                                                                                                                                                                                                                                                                                                                                                                                                                                                                                                                                                                                                                                                                                                                                                                                                                                                                                                                                                                                                                                                                                                                                                                                                                                                                                                                                                                                            \\ \hline
38 & sentiment-no\_definition-score-explanation             & I want you to perform a data annotation task. Your task is to carefully read the text and identify the polarity of the sentiment that is conveyed. Your response must belong to one of the five classes, depending on whether the text is very positive, somewhat positive, neutral, somewhat negative, or very negative. In your output, I want you to provide a probability score for each of the 5 classes. The probability of each class should be a number between 0 and 1 where higher numbers represent a higher probability of that class. Since there are only 5 possible classes, the sum of their probability scores should always be equal to 1. For each class, respond with the name of the class followed by its probability score in each line. In your output, I also want you to provide an explanation for the output. Provide your response in the first line and provide the explanation for your response in the second line. Text: \textless{}\textgreater{}                                                                                                                                                                                                                                                                                                                                                                                                                                                                                                                                                                                                                                                                                                                                                                                                                                                                                                                                                                                                                                                                                                                                                                                                                                                                                                                                                                                                     \\ \hline
39 & sentiment-no\_definition-score-explanation-concise     & Perform a data annotation task by identifying the sentiment polarity of a given text. Choose from five classes: very positive, somewhat positive, neutral, somewhat negative, or very negative. Provide a probability score for each class, ranging from 0 to 1, with the sum of all scores equaling 1. Output the name of each class followed by its probability score, along with an explanation for your response. Text: \textless{}\textgreater{}                                                                                                                                                                                                                                                                                                                                                                                                                                                                                                                                                                                                                                                                                                                                                                                                                                                                                                                                                                                                                                                                                                                                                                                                                                                                                                                                                                                                                                                                                                                                                                                                                                                                                                                                                                                                                                                                                                                                   \\ \hline
40 & frames-no\_definition-label-no\_explanation            & I want you to perform a data annotation task. Your task is to carefully read the headline of a news article and determine the frame(s) of the news article. Each news headline must be assigned one or more of the following 9 frame classes: Politics, Public opinion, Society and culture, Economic consequences, Gun rights, Gun control, Mental health, School and public space safety, Race and ethnicity. In your output, respond with the frame class the headline belongs to. In your response, you may provide one additional class if you believe the headline belongs to multiple classes. Do not respond with more than 2 classes. Only respond with the name of the classes. Do not respond with any other output. Do not provide any other outputs or any explanation for your output. Headline: \textless{}\textgreater{}                                                                                                                                                                                                                                                                                                                                                                                                                                                                                                                                                                                                                                                                                                                                                                                                                                                                                                                                                                                                                                                                                                                                                                                                                                                                                                                                                                                                                                                                                                                                                \\ \hline
41 & frames-no\_definition-label-no\_explanation-concise    & Perform data annotation by assigning news headlines to one or more of 9 frame classes: Politics, Public opinion, Society and culture, Economic consequences, Gun rights, Gun control, Mental health, School and public space safety, Race and ethnicity. Provide only the name of the class(es) without any additional output or explanation. Do not assign more than 2 classes per headline. Headline: \textless{}\textgreater{}                                                                                                                                                                                                                                                                                                                                                                                                                                                                                                                                                                                                                                                                                                                                                                                                                                                                                                                                                                                                                                                                                                                                                                                                                                                                                                                                                                                                                                                                                                                                                                                                                                                                                                                                                                                                                                                                                                                                                       \\ \hline
42 & frames-no\_definition-label-explanation                & I want you to perform a data annotation task. Your task is to carefully read the headline of a news article and determine the frame(s) of the news article. Each news headline must be assigned one or more of the following 9 frame classes: Politics, Public opinion, Society and culture, Economic consequences, Gun rights, Gun control, Mental health, School and public space safety, Race and ethnicity. In your output, respond with the frame class the headline belongs to. In your response, you may provide one additional class if you believe the headline belongs to multiple classes. Do not respond with more than 2 classes. Only respond with the name of the classes. Do not respond with any other output. In your output, I also want you to provide an explanation for the output. Provide your response in the first line and provide the explanation for your response in the second line. Headline: \textless{}\textgreater{}                                                                                                                                                                                                                                                                                                                                                                                                                                                                                                                                                                                                                                                                                                                                                                                                                                                                                                                                                                                                                                                                                                                                                                                                                                                                                                                                                                                                                                 \\ \hline
43 & frames-no\_definition-label-explanation-concise        & Perform a data annotation task by assigning news headlines to one or more of 9 frame classes: Politics, Public opinion, Society and culture, Economic consequences, Gun rights, Gun control, Mental health, School and public space safety, Race and ethnicity. Provide only one or two class names in your output and an explanation for your choice. Headline: \textless{}\textgreater{}                                                                                                                                                                                                                                                                                                                                                                                                                                                                                                                                                                                                                                                                                                                                                                                                                                                                                                                                                                                                                                                                                                                                                                                                                                                                                                                                                                                                                                                                                                                                                                                                                                                                                                                                                                                                                                                                                                                                                                                              \\ \hline
44 & frames-no\_definition-score-no\_explanation            & I want you to perform a data annotation task. Your task is to carefully read the headline of a news article and determine the frame(s) of the news article. Each news headline must be assigned one or more of the following 9 frame classes: Politics, Public opinion, Society and culture, Economic consequences, Gun rights, Gun control, Mental health, School and public space safety, Race and ethnicity. In your output, I want you to provide a probabilistic response for each of the 9 frame classes. The probability of each class should be a number between 0 and 1 where higher numbers represent a higher probability of that class. Since there are 9 possible classes, the sum of their probability scores should always be equal to 1. For each class, respond with the name of the class followed by its probability score in each line. Do not provide any other outputs or any explanation for your output. Headline: \textless{}\textgreater{}                                                                                                                                                                                                                                                                                                                                                                                                                                                                                                                                                                                                                                                                                                                                                                                                                                                                                                                                                                                                                                                                                                                                                                                                                                                                                                                                                                                                                    \\ \hline
45 & frames-no\_definition-score-no\_explanation-concise    & Perform data annotation by assigning news headlines to 1 or more of 9 frame classes: Politics, Public opinion, Society and culture, Economic consequences, Gun rights, Gun control, Mental health, School and public space safety, Race and ethnicity. Provide a probabilistic response for each class, with a score between 0 and 1. The sum of scores for all classes should be 1. Output only the class name and its probability score for each line. No additional output or explanation needed. Headline: \textless{}\textgreater{}                                                                                                                                                                                                                                                                                                                                                                                                                                                                                                                                                                                                                                                                                                                                                                                                                                                                                                                                                                                                                                                                                                                                                                                                                                                                                                                                                                                                                                                                                                                                                                                                                                                                                                                                                                                                                                                \\ \hline
46 & frames-no\_definition-score-explanation                & I want you to perform a data annotation task. Your task is to carefully read the headline of a news article and determine the frame(s) of the news article. Each news headline must be assigned one or more of the following 9 frame classes: Politics, Public opinion, Society and culture, Economic consequences, Gun rights, Gun control, Mental health, School and public space safety, Race and ethnicity. In your output, I want you to provide a probabilistic response for each of the 9 frame classes. The probability of each class should be a number between 0 and 1 where higher numbers represent a higher probability of that class. Since there are 9 possible classes, the sum of their probability scores should always be equal to 1. For each class, respond with the name of the class followed by its probability score in each line. In your output, I also want you to provide an explanation for the output. Provide your response in the first line and provide the explanation for your response in the second line. Headline: \textless{}\textgreater{}                                                                                                                                                                                                                                                                                                                                                                                                                                                                                                                                                                                                                                                                                                                                                                                                                                                                                                                                                                                                                                                                                                                                                                                                                                                                                                     \\ \hline
47 & frames-no\_definition-score-explanation-concise        & Perform data annotation by assigning news article frames. Read headlines and assign one or more of 9 frame classes: Politics, Public opinion, Society and culture, Economic consequences, Gun rights, Gun control, Mental health, School and public space safety, Race and ethnicity. Output probabilistic response for each class, with a number between 0 and 1. Sum of probability scores should equal 1. Provide name of class and probability score for each line. Explain output in first and second line of response. Headline: \textless{}\textgreater{}                                                                                                                                                                                                                                                                                                                                                                                                                                                                                                                                                                                                                                                                                                                                                                                                                                                                                                                                                                                                                                                                                                                                                                                                                                                                                                                                                                                                                                                                                                                                                                                                                                                                                                                                                                                                                        \\ \hline
48 & frames-definition-label-no\_explanation                & I want you to perform a data annotation task. Your task is to carefully read the headline of a news article and determine the frame(s) of the news article. Each news headline must be assigned one or more of the following 9 frame classes: Politics, Public opinion, Society and culture, Economic consequences, Gun rights, Gun control, Mental health, School and public space safety, Race and ethnicity. Annotation guidelines: For this task, additional instructions for each of the frame class are provided below: 1) Gun rights: The story is related to the Constitution, the second amendment, and protection of individual liberty and gun ownership as a right, 2) Gun control: The story is about issues related to regulating guns through legislation and other institutional measures. 3) Politics: The story is mainly about the political issues around guns and shootings. 4) Mental health: The story is about issues related to individuals' mental illnesses or emotional well-being, or the mental health system as a whole. 5) School and public space safety: Issues related to institutional and school safety 6) Race and ethnicity: The story is about gun issues related to certain ethnic group(s) 7) Public opinion: The story is about the public's, including a certain community's reactions to gun-related issues. 8) Society and culture: Societal-wide factors that are related to gun violence. 9) Economic consequences: The story is about financial losses or gains, or the costs involved in gun-related issues.  You must follow the instructions mentioned above when providing your response. Do not provide a response that does not align with the instructions. In your output, respond with the frame class the headline belongs to. In your response, you may provide one additional class if you believe the headline belongs to multiple classes. Do not respond with more than 2 classes. Only respond with the name of the classes. Do not respond with any other output. Do not provide any other outputs or any explanation for your output. Headline: \textless{}\textgreater{}                                                                                                                                                                                                                                            \\ \hline
49 & frames-definition-label-no\_explanation-concise        & Perform data annotation by assigning news headlines to one or more of 9 frame classes: Politics, Public opinion, Society and culture, Economic consequences, Gun rights, Gun control, Mental health, School and public space safety, Race and ethnicity. Follow provided guidelines for each class. 1) Gun rights: The story is related to the Constitution, the second amendment, and protection of individual liberty and gun ownership as a right, 2) Gun control: The story is about issues related to regulating guns through legislation and other institutional measures. 3) Politics: The story is mainly about the political issues around guns and shootings. 4) Mental health: The story is about issues related to individuals' mental illnesses or emotional well-being, or the mental health system as a whole. 5) School and public space safety: Issues related to institutional and school safety 6) Race and ethnicity: The story is about gun issues related to certain ethnic group(s) 7) Public opinion: The story is about the public's, including a certain community's reactions to gun-related issues. 8) Society and culture: Societal-wide factors that are related to gun violence. 9) Economic consequences: The story is about financial losses or gains, or the costs involved in gun-related issues.  Provide only the name of the class(es) and do not exceed 2 classes. Headline: \textless{}\textgreater{}                                                                                                                                                                                                                                                                                                                                                                                                                                                                                                                                                                                                                                                                                                                                                                                                                                                                                                                                           \\ \hline
50 & frames-definition-label-explanation                    & I want you to perform a data annotation task. Your task is to carefully read the headline of a news article and determine the frame(s) of the news article. Each news headline must be assigned one or more of the following 9 frame classes: Politics, Public opinion, Society and culture, Economic consequences, Gun rights, Gun control, Mental health, School and public space safety, Race and ethnicity. Annotation guidelines: For this task, additional instructions for each of the frame class are provided below: 1) Gun rights: The story is related to the Constitution, the second amendment, and protection of individual liberty and gun ownership as a right, 2) Gun control: The story is about issues related to regulating guns through legislation and other institutional measures. 3) Politics: The story is mainly about the political issues around guns and shootings. 4) Mental health: The story is about issues related to individuals' mental illnesses or emotional well-being, or the mental health system as a whole. 5) School and public space safety: Issues related to institutional and school safety 6) Race and ethnicity: The story is about gun issues related to certain ethnic group(s) 7) Public opinion: The story is about the public's, including a certain community's reactions to gun-related issues. 8) Society and culture: Societal-wide factors that are related to gun violence. 9) Economic consequences: The story is about financial losses or gains, or the costs involved in gun-related issues.  You must follow the instructions mentioned above when providing your response. Do not provide a response that does not align with the instructions. In your output, respond with the frame class the headline belongs to. In your response, you may provide one additional class if you believe the headline belongs to multiple classes. Do not respond with more than 2 classes. Only respond with the name of the classes. Do not respond with any other output. In your output, I also want you to provide an explanation for the output. Provide your response in the first line and provide the explanation for your response in the second line. Headline: \textless{}\textgreater{}                                                                                                                             \\ \hline
51 & frames-definition-label-explanation-concise            & Perform data annotation by assigning news article frames. Assign one or more of the following 9 frame classes to each headline: Politics, Public opinion, Society and culture, Economic consequences, Gun rights, Gun control, Mental health, School and public space safety, Race and ethnicity. Follow the provided guidelines for each class. 1) Gun rights: The story is related to the Constitution, the second amendment, and protection of individual liberty and gun ownership as a right, 2) Gun control: The story is about issues related to regulating guns through legislation and other institutional measures. 3) Politics: The story is mainly about the political issues around guns and shootings. 4) Mental health: The story is about issues related to individuals' mental illnesses or emotional well-being, or the mental health system as a whole. 5) School and public space safety: Issues related to institutional and school safety 6) Race and ethnicity: The story is about gun issues related to certain ethnic group(s) 7) Public opinion: The story is about the public's, including a certain community's reactions to gun-related issues. 8) Society and culture: Societal-wide factors that are related to gun violence. 9) Economic consequences: The story is about financial losses or gains, or the costs involved in gun-related issues.  Provide only one or two classes in your response and an explanation for your choice. Do not provide any other output. Headline: \textless{}\textgreater{}                                                                                                                                                                                                                                                                                                                                                                                                                                                                                                                                                                                                                                                                                                                                                                                                                                            \\ \hline
52 & frames-definition-score-no\_explanation                & I want you to perform a data annotation task. Your task is to carefully read the headline of a news article and determine the frame(s) of the news article. Each news headline must be assigned one or more of the following 9 frame classes: Politics, Public opinion, Society and culture, Economic consequences, Gun rights, Gun control, Mental health, School and public space safety, Race and ethnicity. Annotation guidelines: For this task, additional instructions for each of the frame class are provided below: 1) Gun rights: The story is related to the Constitution, the second amendment, and protection of individual liberty and gun ownership as a right, 2) Gun control: The story is about issues related to regulating guns through legislation and other institutional measures. 3) Politics: The story is mainly about the political issues around guns and shootings. 4) Mental health: The story is about issues related to individuals' mental illnesses or emotional well-being, or the mental health system as a whole. 5) School and public space safety: Issues related to institutional and school safety 6) Race and ethnicity: The story is about gun issues related to certain ethnic group(s) 7) Public opinion: The story is about the public's, including a certain community's reactions to gun-related issues. 8) Society and culture: Societal-wide factors that are related to gun violence. 9) Economic consequences: The story is about financial losses or gains, or the costs involved in gun-related issues.  You must follow the instructions mentioned above when providing your response. Do not provide a response that does not align with the instructions. In your output, I want you to provide a probabilistic response for each of the 9 frame classes. The probability of each class should be a number between 0 and 1 where higher numbers represent a higher probability of that class. Since there are 9 possible classes, the sum of their probability scores should always be equal to 1. For each class, respond with the name of the class followed by its probability score in each line. Do not provide any other outputs or any explanation for your output. Headline: \textless{}\textgreater{}                                                                                                                \\ \hline
53 & frames-definition-score-no\_explanation-concise        & Perform data annotation by assigning news headlines to one or more of the 9 frame classes: Politics, Public opinion, Society and culture, Economic consequences, Gun rights, Gun control, Mental health, School and public space safety, Race and ethnicity. Follow the provided guidelines for each class. 1) Gun rights: The story is related to the Constitution, the second amendment, and protection of individual liberty and gun ownership as a right, 2) Gun control: The story is about issues related to regulating guns through legislation and other institutional measures. 3) Politics: The story is mainly about the political issues around guns and shootings. 4) Mental health: The story is about issues related to individuals' mental illnesses or emotional well-being, or the mental health system as a whole. 5) School and public space safety: Issues related to institutional and school safety 6) Race and ethnicity: The story is about gun issues related to certain ethnic group(s) 7) Public opinion: The story is about the public's, including a certain community's reactions to gun-related issues. 8) Society and culture: Societal-wide factors that are related to gun violence. 9) Economic consequences: The story is about financial losses or gains, or the costs involved in gun-related issues.  Provide a probabilistic response for each class, with a number between 0 and 1 representing the probability of that class. The sum of all probabilities should be 1. Output the name of the class followed by its probability score for each line. No other output or explanation is required. Headline: \textless{}\textgreater{}                                                                                                                                                                                                                                                                                                                                                                                                                                                                                                                                                                                                                                                                                                        \\ \hline
54 & frames-definition-score-explanation                    & I want you to perform a data annotation task. Your task is to carefully read the headline of a news article and determine the frame(s) of the news article. Each news headline must be assigned one or more of the following 9 frame classes: Politics, Public opinion, Society and culture, Economic consequences, Gun rights, Gun control, Mental health, School and public space safety, Race and ethnicity. Annotation guidelines: For this task, additional instructions for each of the frame class are provided below: 1) Gun rights: The story is related to the Constitution, the second amendment, and protection of individual liberty and gun ownership as a right, 2) Gun control: The story is about issues related to regulating guns through legislation and other institutional measures. 3) Politics: The story is mainly about the political issues around guns and shootings. 4) Mental health: The story is about issues related to individuals' mental illnesses or emotional well-being, or the mental health system as a whole. 5) School and public space safety: Issues related to institutional and school safety 6) Race and ethnicity: The story is about gun issues related to certain ethnic group(s) 7) Public opinion: The story is about the public's, including a certain community's reactions to gun-related issues. 8) Society and culture: Societal-wide factors that are related to gun violence. 9) Economic consequences: The story is about financial losses or gains, or the costs involved in gun-related issues.  You must follow the instructions mentioned above when providing your response. Do not provide a response that does not align with the instructions. In your output, I want you to provide a probabilistic response for each of the 9 frame classes. The probability of each class should be a number between 0 and 1 where higher numbers represent a higher probability of that class. Since there are 9 possible classes, the sum of their probability scores should always be equal to 1. For each class, respond with the name of the class followed by its probability score in each line. In your output, I also want you to provide an explanation for the output. Provide your response in the first line and provide the explanation for your response in the second line. Headline: \textless{}\textgreater{} \\ \hline
55 & frames-definition-score-explanation-concise            & Perform data annotation by assigning news headlines to one or more of the 9 frame classes: Politics, Public opinion, Society and culture, Economic consequences, Gun rights, Gun control, Mental health, School and public space safety, Race and ethnicity. Follow the provided guidelines for each class. 1) Gun rights: The story is related to the Constitution, the second amendment, and protection of individual liberty and gun ownership as a right, 2) Gun control: The story is about issues related to regulating guns through legislation and other institutional measures. 3) Politics: The story is mainly about the political issues around guns and shootings. 4) Mental health: The story is about issues related to individuals' mental illnesses or emotional well-being, or the mental health system as a whole. 5) School and public space safety: Issues related to institutional and school safety 6) Race and ethnicity: The story is about gun issues related to certain ethnic group(s) 7) Public opinion: The story is about the public's, including a certain community's reactions to gun-related issues. 8) Society and culture: Societal-wide factors that are related to gun violence. 9) Economic consequences: The story is about financial losses or gains, or the costs involved in gun-related issues.  Provide a probabilistic response for each class, with a number between 0 and 1 representing the probability of that class. The sum of all probabilities should be 1. Output the name of the class followed by its probability score in each line. Provide an explanation for the output. Headline: \textless{}\textgreater{}                                                                                                                                                                                                                                                                                                                                                                                                                                                                                                                                                                                                                                                                                                              \\ \hline

\end{longtable}

\end{document}